\documentclass[journal]{IEEEtran}
\IEEEoverridecommandlockouts
\usepackage{cite}
\usepackage{cuted}
\usepackage{soul}
\usepackage{bm}
\usepackage{amsmath,amssymb,amsfonts}
\usepackage{lipsum}
\usepackage{graphicx} 
  
\ifCLASSOPTIONcompsoc
\usepackage[caption=false,font=normalsize,labelfont=sf,textfont=sf]{subfig}
\else
\usepackage[caption=false,font=footnotesize]{subfig}
\fi
\usepackage{textcomp}
\usepackage{color}
\usepackage{xcolor}
\usepackage{nomencl}
\usepackage{url}
\usepackage{subfig}
\makenomenclature
\usepackage{multirow}
\usepackage{array}
\usepackage{makecell}
\usepackage{booktabs}
\usepackage{tabularx}

\usepackage{algorithm}  
\usepackage{algorithmic}

\usepackage{todo}
\def\BibTeX{{\rm B\kern-.05em{\sc i\kern-.025em b}\kern-.08em
    T\kern-.1667em\lower.7ex\hbox{E}\kern-.125emX}}

\usepackage{fancyhdr}

\begin{document}
\bstctlcite{BSTcontrol}

\definecolor{limegreen}{rgb}{0.2, 0.8, 0.2}
\definecolor{forestgreen}{rgb}{0.13, 0.55, 0.13}
\definecolor{greenhtml}{rgb}{0.0, 0.5, 0.0}

\title{Fault Diagnosis and Quantification for Photovoltaic Arrays based on Differentiable Physical Models}

\author{
	\vskip 1em	
	Zenan Yang, \emph{Student Membership},
	Yuanliang Li,
	Jingwei Zhang,
    \\Yongjie Liu, \emph{Student Membership},
    and Kun Ding
	\thanks{
        Zenan Yang and Yuanliang Li contributed equally to this work.
		
		Zenan Yang and Jingwei Zhang are with the College of Mechanical and Electrical Engineering, Hohai University, Changzhou, 213200, China (e-mail: zenanyang@hhu.edu.cn; jwzhang@hhu.edu.cn). 

        Yuanliang Li is with the Concordia Institute for Information Systems Engineering, Concordia University, Montr\'eal, H3G1M8, Canada (l\_yuanli@live.concordia.ca).

        Yongjie Liu is with the AAU Energy, Aalborg University, Aalborg, 9220, Denmark (yoli@energy.aau.dk).
		
		Kun Ding is the corresponding author with the College of Mechanical and Electrical Engineering, Hohai University, Changzhou, 213200, China (e-mail: dingk@hhu.edu.cn).

        \textbf{\textit{This work has been submitted to the IEEE for possible publication. Copyright may be transferred without notice, after which this version may no longer be accessible.}}
	}
}

\maketitle

\begin{abstract}
Accurate fault diagnosis and quantification are essential for the reliable operation and intelligent maintenance of photovoltaic (PV) arrays. However, existing fault quantification methods often suffer from limited efficiency and interpretability. To address these challenges, this paper proposes a novel fault quantification approach for PV strings based on a differentiable fast fault simulation model (DFFSM). The proposed DFFSM accurately models I-V characteristics under multiple faults and provides analytical gradients with respect to fault parameters. Leveraging this property, a gradient-based fault parameters identification (GFPI) method using the Adahessian optimizer is developed to efficiently quantify partial shading, short-circuit, and series-resistance degradation. Experimental results on both simulated and measured I-V curves demonstrate that the proposed GFPI achieves high quantification accuracy across different faults, with the I-V reconstruction error below 3\%, confirming the feasibility and effectiveness of the application of differentiable physical simulators for PV system fault diagnosis.
\end{abstract}

\begin{IEEEkeywords}
Photovoltaic system, fault modeling, fault quantification, differentiable physical model, Adahessian optimizer
\end{IEEEkeywords}

\section{Introduction}
Driven by the global pursuit of carbon neutrality, solar photovoltaic (PV) is becoming the dominate renewable energy resource \cite{IEA2025}. As PV arrays constitute a fundamental part of the energy infrastructure, maintaining their reliable and efficient operation is essential. However, since PV arrays operate under harsh outdoor conditions, they are prone to various types of fault, such as partial shading (PS), soiling, potential-induced degradation (PID), short-circuit (SC), open-circuit (OC), line-line (LL) fault, line-ground (LG) fault, among others, which can reduce their energy yield and shorten system lifetime \cite{Kumar2021Identification}. Thus, effective health monitoring and fault diagnosis techniques are indispensable for PV systems. 

Data sources used in fault diagnosis for PV systems include operational electrical measurements (voltage and current), I-V characteristic curves obtained via I-V scanning, electroluminescence (EL) images, and infrared (IR) thermal images, among others \cite{Yang2024Recent}. Various diagnostic methods have been developed for each data type. For image data, deep neural networks (DNNs), e.g., CNN and YOLO, are widely applied to detect cell cracks, hot-spots, and soiling in PV panels \cite{Li2021Application}. However, image-based methods typically operate offline, lack real-time capability, and require expensive imaging equipments. By contrast, PV array voltage and current measurements can be readily obtained from inverters, enabling real-time diagnosis without high-cost external sensors. Moreover, many modern string PV inverters now integrate online I-V scanning functionality, which directly measures the I-V curves of PV strings \cite{Pillai2019Comparative}. This provides rich fault-related information that supports advanced fault diagnosis. Therefore, I-V curves are the primary data utilized in this study.

Fault diagnosis research generally follows two directions: fault classification and fault quantification. Fault classification determines the type of a fault, in which many methods have been proposed. For example, Ma et al. simply applied the slop of the I-V curves along with a threshold-based method to recognize PS, hot-spots, and cell cracks \cite{Ma2021Photovoltaic}. 
With advances in artificial intelligence (AI) techniques, machine-learning (ML)-based classifiers built on explicitly extracted features from I-V curves have been widely studied. For instance, Eskandari et al. defined 16 normalized features from I-V curves using short-circuit current, open-circuit voltage, maximum power point voltage, and trained a set of hierarchical ML models to detect LL and LG faults \cite{Eskandari2021Fault}. 
Many recent methods can directly process I-V curves without explicit and manual feature extractions.
Liu et al. trained contrastive learning models to automatically extract features from I-V curves and then adopted Siamese networks to diagnose PS, SC, and OC \cite{Liu2025SelfCorrectingGuided}.
Ha et al. developed a densely connected convolutional network (DenseNet) trained directly on I-V curves to classify LL fault, OC, PS, and degradation \cite{Ha2024DataDriven}. 
Ren et al. adopted a similarity learning method using a Siamese-MobileNet (SiamMN) network trained on randomly paired I-V curves, which works well with limited labeled data and enables accurate diagnosis of common PV faults under diverse conditions \cite{Ren2025Fault}.

However, fault classification methods remain primarily qualitative and do not provide detailed descriptions about the faults. In contrast, fault quantification provides numerical indicators of fault characteristics or severity, thereby enabling advanced applications such as intelligent operation and maintenance (O\&M) and prognostics and health management (PHM) \cite{Chang2024Prognostics}. 

Many fault quantification approaches have been proposed. 
Bastidas-Rodríguez et al. quantified the degradation by estimating the increases in the series resistance based on the deviation between the measured maximum power point and the model-predicted value under healthy conditions \cite{Bastidas2015ModelBased}. 
Zhang et al. developed an I-V curve reconstruction method to quantitatively estimate the number of shaded modules and the severity of shading \cite{zhang2024v}.
He et al. proposed a method that converts I-V curves to images and feeds them into a two-stream DNN, enabling both the detection and severity quantification of five common shadings in PV arrays \cite{He2025Shading}. However, these methods generally quantify only a single type of fault, while in reality multiple fault types, such as PS and degradation, may occur simultaneously. To handle this limitation, Li et al. developed a numerical simulator for PV strings and applied the differential evolution (DE) algorithm to identify fault parameters for PS, SC, and series-resistance degradation (SRD), by minimizing the discrepancy between measured and simulated I-V curves \cite{Li2019Fault}. Nevertheless, meta-heuristic optimization methods require large population sizes to sufficiently explore the fault parameter space, leading to considerable computational and memory costs. In addition, these methods treat the PV simulator as a black box, relying solely on its output to guide the search. As a result, they ignore valuable physics-related information, which in turn limits their fault quantification efficiency and interpretability.

To address the above limitations, we develop a differentiable fast fault simulation model (DFFSM) trailed for PV string fault modeling and quantification. DFFSM is a differentiable physical simulator that not only models I-V characteristics of PV strings under multiple faults, but also provides fault-related gradients. This enables physics-informed optimization to identify fault parameters directly, rather than relying on heuristic search. We apply Adahessian as a gradient-based optimizer and employ projected gradient decent (PGD) to identify fault parameters that can quantify PS, SC, and SRD. As a result, the proposed solution offers high computational efficiency, low memory cost, and strong physical interpretability.

Moreover, DFFSM's differentiability allows the use of mainstream automatic differentiation frameworks (e.g. PyTorch and TensorFlow), enabling seamless deployment. Furthermore, its low computational and memory costs also allows deployment on edge devices, such as PV inverters and I-V scanners, as an auxiliary feature. Finally, DFFSM can support the training of both upstream and downstream AI models with accelerated convergence, as the gradients can propagate through the AI pipelines, facilitating end-to-end applications.

The main contributions of the paper are as following:
\begin{itemize}
\item To the best of our knowledge, this is the first work to develop a differentiable physical simulator (i.e., DFFSM) for PV fault modeling and quantification. Its forward calculation estimates the I-V curve under multiple faults, and its backward calculation provides analytical gradients with respect to the fault parameters.
\item A gradient-based fault parameters identification method (GFPI) based on the DFFSM and Adahessian optimizer is proposed to quantify PS, SC, and SRD, with high computational efficiency and strong physical interpretability. 
\item A comparative evaluation of gradient-based optimizers demonstrates that Adahessian achieves superior efficiency in fault quantification. Validation on measured I-V curves further confirms the accuracy and effectiveness of the proposed GFPI in quantifying PS, SC, and SRD.
\item The implementation of the DFFSM and GFPI is released as open-source to encourage further innovation in differentiable physical simulators and physics-informed fault diagnosis for PV systems, which is available at \url{https://github.com/Yuanliang-Li/dffsm}.
\end{itemize}

The rest of the paper is organized as follows. Section \ref{section-dffsm} introduces the DFFSM. Section \ref{GradientFPI} explains the GFPI method. Section \ref{Exp} presents the experimental validation. Section \ref{Conclusion} concludes the paper.

\begin{figure}[h]
\centering
\includegraphics[width=0.45\textwidth]{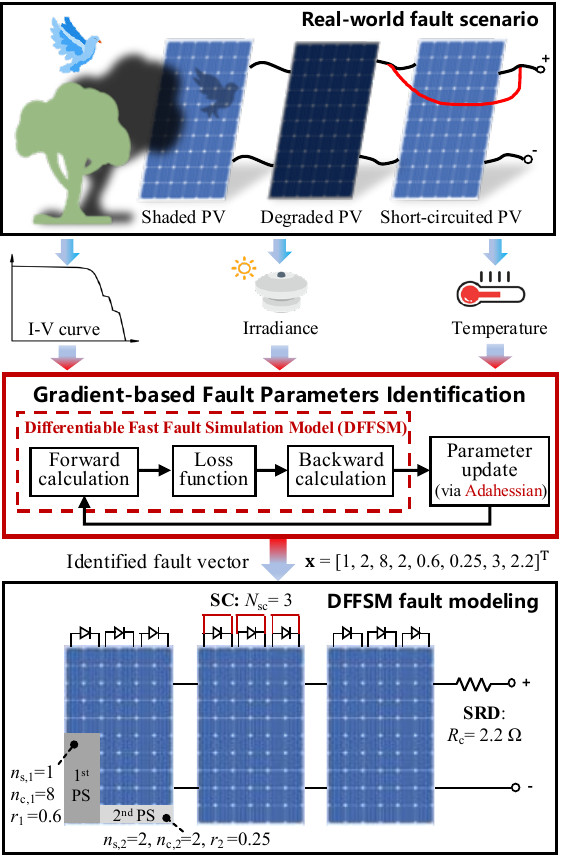}
\caption{Framework of the proposed fault quantification method.}
\label{fig:Framework}
\end{figure}

\section{Differentiable Fast Fault Simulation Model}\label{section-dffsm}
Fig.~\ref{fig:Framework} illustrates the proposed fault quantification framework for PV strings. Given a measured I-V curve of a string, along with its in-plane irradiance and cell temperature, our framework performs gradient-based fault parameter identification (GFPI), which outputs a fault vector that quantifies the faults present in the string.

A core component of GFPI is the DFFSM, a differentiable PV simulator that performs efficient forward calculation, loss calculation, and backward calculations. The forward calculation implements the physical model of a PV string based on the reverse-biased single diode model (RSDM), which estimates the output characteristics of the string (I-V curve) under identified fault parameters. The loss function determines the identification error (loss). The backward calculation is the key process that makes the DFFSM differentiable, which calculates the gradient of a defined loss function with respect to each fault parameter. GFPI applies Adahessian optimizer to update the fault parameters iteratively. 

This section provides the details of DFFSM, including the forward calculation, loss function, and backward calculation.

\subsection{Reverse-biased Single Diode Model}
The proposed DFFSM is based on the principle of voltage superposition of all PV cells in a PV string. For the PV cell model, it adopts the RSDM as the basic physical model, which improves the single diode model (SDM) by accounting for the reverse leakage current of the cell's PN junction under negative voltage. String models based on RSDM can provide an accurate representation under partial shading since shaded PV cells can operate at negative voltages \cite{Clement2021Illumination}.

\vspace{-10pt}
\begin{figure}[h]
\centering
\includegraphics[width=0.7\columnwidth]{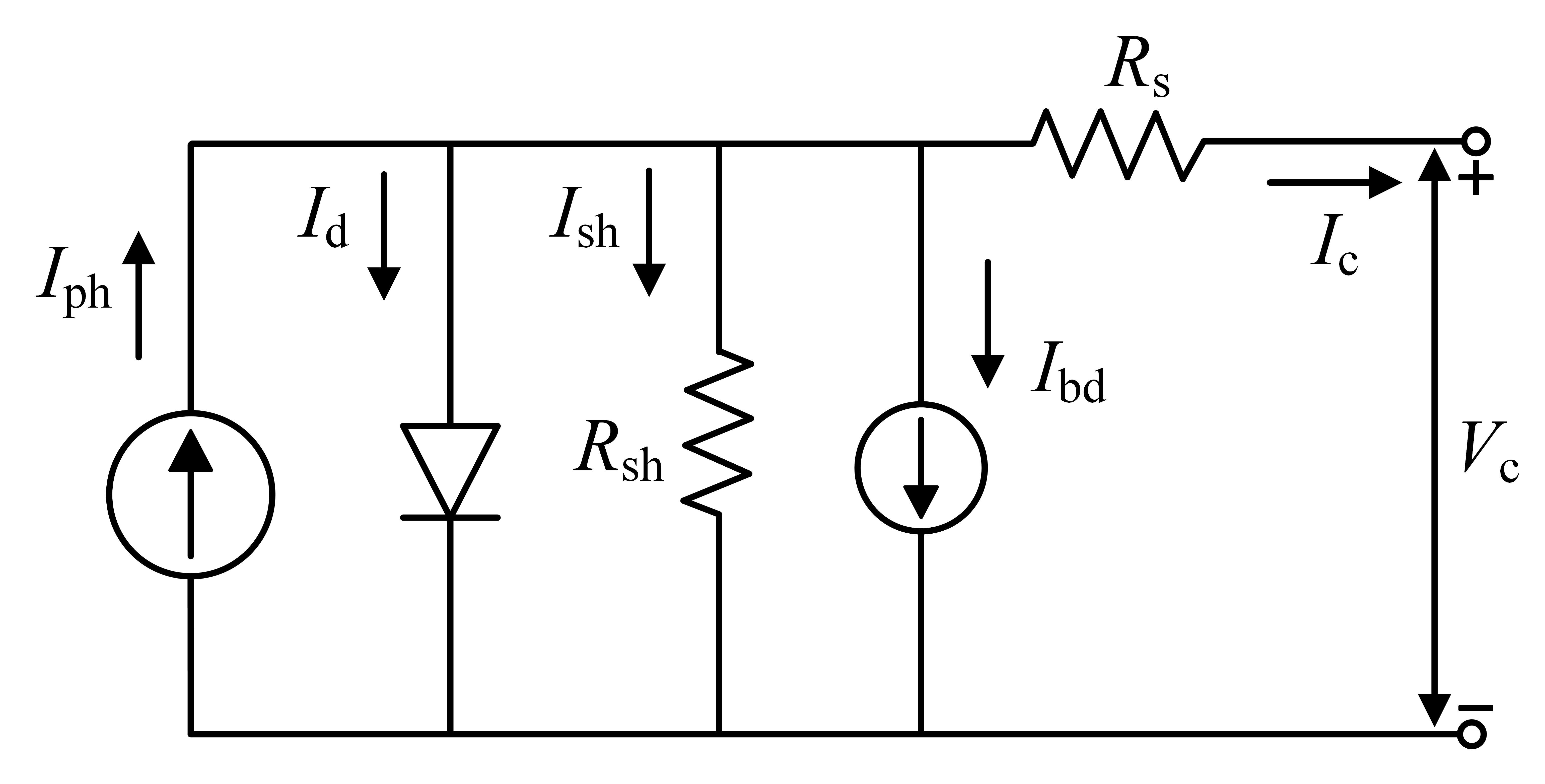}
\caption{Electrical equivalent circuit of PV cells based on the RSDM.}
\label{fig:RSDM}
\end{figure}

Fig.~\ref{fig:RSDM} illustrates the electrical equivalent circuit of PV cells based on the RSDM. The output I-V characteristic is expressed as follows:
\begin{equation}\label{eq-rsdm}
\begin{split}
    I_{\rm c}  = I_{\rm ph} &- 
    \underbrace{I_0 \left[ \exp \left( \frac{V_{\rm c}+I_{\rm c} R_{\rm s}}{n V_{\rm th}} \right) - 1 \right]}_{I_{\rm d}} 
    - \underbrace{\frac{V_{\rm c} + I_{\rm c} R_{\rm s}}{R_{\rm sh}}}_{I_{\rm sh}} \\
    &- \underbrace {a \frac{V_{\rm c} + I_{\rm c} R_{\rm s}}{R_{\rm sh}}\left( 1-\frac{V_{\rm c}+I_{\rm c} R_{\rm s}}{V_{\rm br}} \right) ^{-m} }_{I_{\rm bd}}
\end{split}
\end{equation}
where $I_{\rm c}$ and $V_{\rm c}$ denote the output current and voltage of the PV cell, respectively. $I_{\rm ph}$ is the photocurrent. $I_0$ is the diode reverse saturation current. $n$ is the diode ideality factor. $R_{\rm s}$ and $R_{\rm sh}$ are the equivalent series and shunt resistances of the PV cell, respectively \cite{Chen2023Explicit}. 
$V_{\rm th}=kT/q$ is the thermal voltage, where $k$ is the boltzmann constant ($1.38 \times 10^{-23}$J/K), $T$ is the cell temperature in Kelvins (K) and  $q$ is the elementary charge ($1.602 \times 10^{-19}$C). $a$, $V_{\rm br}$, and $m$ denote the fraction of ohmic current involved in avalanche breakdown, the breakdown voltage, and the avalanche breakdown exponent, respectively,  which are set to $0.002$, $-21.29\text{V}$, and 3 \cite{Liu2020Improved}. The five components, $I_{\rm ph}$, $I_0$, $n$, $R_{\rm s}$, and $R_{\rm sh}$, are computed using the De Soto five‑parameter model (CEC performance model) in \cite{DeSoto2006Improvement}. Each component is expressed as a function of its value under standard test condition (STC) ($I_{\rm ph,stc}$, $I_{\rm 0,stc}$, $n_{\rm stc}$, $R_{\rm s,stc}$, $R_{\rm sh,stc}$), together with the in-plane irradiance ($G$, unit: $\text{W/m}^2$) and cell temperature ($T$, unit: K).

$I_{\rm ph,stc}$, $I_{\rm 0,stc}$, $n_{\rm stc}$, $R_{\rm s,stc}$, and $R_{\rm sh,stc}$ are five model parameters of the RSDM, which should be pre-determined before applying the DFFSM. These parameters can be extracted from measured I-V curves of PV strings under normal operating conditions and should be updated regularly, as they vary with seasonal changes and the inherent aging of PV modules. 
Various parameter extraction techniques can be employed, including meta-heuristic optimization-based methods \cite{Toledo2018TwoStep, Subudhi2018Bacterial, Li2021Comprehensive, Kharchouf2022Parameters} and reinforcement learning-based methods \cite{Zhang2023Neural, Wang2025Reinforcement}. 

\subsection{Fault Vector}\label{FaultVector}
During the long-term operation, PV arrays are prone to various types of fault. This paper specifically focuses on three types of fault: PS, SC, and SRD. To quantify PS, we adopt the basic idea that a complex shadow on a PV string can be represented by a number of rectangular shadows \cite{Li2019Fault}. We assume there are at most $N_{\rm ps}$ rectangular shadows (control the resolution of PS), each with a distinct shading ratio and a size. As illustrated in Fig.~\ref{fig:Framework}, the size of each shadow can be described by the number of affected substrings ($n_{\rm s}$) and the number of shading cells per substring ($n_{\rm c}$) under this shadow. Each shadow has its own shading ratio ($r$), representing the fraction of irradiance blocked by this shadow.  
The SC fault is quantified by the number of short-circuited bypass diodes in the string ($N_{\rm sc}$). The SRD is quantified by the increased series resistance of the string ($R_{\rm c}$). Finally, a fault vector can be constructed as follows:
\begin{equation}\label{faultvector}
    \mathbf{x} = [\mathbf{n_s}, \mathbf{n_c}, \mathbf{r}, N_{\rm sc}, R_{\rm c}]^{\top} 
\end{equation}
where $\mathbf{n_s}=[n_{\rm s,1},n_{\rm s,2},\cdots,n_{{\rm s},N_{\rm ps}}]$ with $n_{{\rm s},i}$ representing the number of affected substring under the $i^{\rm th}$ shadow, $\mathbf{n_c}=[n_{\rm c,1},n_{\rm c,2},\cdots,n_{{\rm c},N_{\rm ps}}]$ with $n_{{\rm c},i}$ representing the number of shading cells per substring under the $i^{\rm th}$ shadow, and $\mathbf{r}=[r_1,r_2,\cdots,r_{N_{\rm ps}}]$ with $r_i$ representing the shading ratio of the $i^{\rm th}$ shadow.

This fault vector serves as the input to the proposed DFFSM to estimate the I-V curve of the PV string. To interpret the fault vector, DFFSM assumes that each substring (defined as a number of PV cells connected in series under a bypass diode) can only be affected by at most one type of fault. For example, one substring may suffer either a PS or a SC. Moreover, because pinpointing the exact location of each fault based solely on a measured I-V curve is challenging, when calculating the I-V curve the DFFSM assigns the first PS shadow to the first substring, with any other faults placed sequentially in the subsequent substrings.

\subsection{Forward Calculation of DFFSM}\label{subsection-forward-propagation}
Given a measured I-V curve of a PV string under known cell temperature, in-plane irradiance, and fault conditions, the forward function of the DFFSM can be expressed as:
\begin{equation}\label{eq-forward}
\mathbf{V} = f(\mathbf{\tilde I}, T, G, \mathbf{x})
\end{equation}
where $\mathbf{\tilde I} \in \mathbb{R}^N$ is the measured current sequence (vector) with $N$ data points, corresponding to the measured voltage sequence $\mathbf{\tilde V} \in \mathbb{R}^N$ of an I-V curve. $T$, $G$, and $\mathbf{x}$ denote the cell temperature (K), in-plane irradiance (W/m$^2$), and fault vector, respectively. The output $\mathbf{V}\in \mathbb{R}^N$ represents the estimated voltage sequence (vector) obtained from the forward function, exhibiting a one-to-one mapping to the $\mathbf{\tilde I}$. 

Algorithm \ref{code:forward-calculation} illustrates the forward calculation procedure and the detailed steps are presented below.

\subsubsection{Construct ambient vectors}
The ambient vectors include a irradiance vector $\mathbf{G}\in\mathbb{R}^{N_{\rm env}}$ and a cell temperature vector $\mathbf{T}\in\mathbb{R}^{N_{\rm env}}$ that cover $N_{\rm env}$ types of environmental conditions of PV cells in the string, where $N_{\rm env} = N_{\rm ps} + 1$ that corresponds to $N_{\rm ps}$ shadows (defined by the fault vector) and one normal condition. Here, we assume all PV cells have the same temperature. Consequently, the ambient vectors can be formulated as $\mathbf{G} = [(1 - r_1)G, \cdots ,(1 - r_{N_{\rm ps}})G, G]^{\top}$, and $\mathbf{T} = [T,\cdots ,T]^{\top}$.

\subsubsection{Calculate the five components of PV cell under each environmental condition}
For each environmental condition in $\mathbf{G}$ and $\mathbf{T}$, we calculate the five components of PV cell, i.e., $I_{\rm ph}$, $I_{\rm 0}$, $n$, $R_{\rm s}$, and $R_{\rm sh}$, using CEC performance model.

\subsubsection{Calculate the output voltages of PV cell for each environmental condition}
The PV cell voltage is calculated by solving \eqref{eq-rsdm} based on Newton iteration for each environmental condition. 
The formula of Newton iteration is presented in \eqref{Vnext}:
\begin{equation}\label{Vnext}
    V_{{\rm c},t+1} = V_{{\rm c},t} - \frac{F\left(V_{{\rm c},t} \right)}{F'\left(V_{{\rm c},t} \right)}
\end{equation}
where $F(V_{\rm c})$ is given as:
\begin{equation}\label{F}
\begin{split}
    F(V_{\rm c}) & = I_{\rm ph}-I_0 \left[ \exp \left( \frac{V_{\rm c}+I_{\rm c} R_{\rm s}}{n V_{\rm th}} \right) - 1 \right] -\frac{V_{\rm c}+I_{\rm c} R_{\rm s}}{R_{\rm sh}} \\ 
    & \left[ 1 + a \left( 1-\frac{V_{\rm c}+I_{\rm c} R_{\rm s}}{V_{\rm br}} \right) ^{-m} \right] - I_{\rm c}
\end{split}
\end{equation}

In this work, the initial guess of the cell voltage ($V_{\rm c,0}$) is calculated using the Lambert W function-based explicit formula for solving the traditional SDM \cite{Xu2022Separable}, shown as follows:
\begin{equation}\label{LambertW}
\begin{split}
    & V_{\rm c,0}^{(i)} = I_{\rm ph}R_{\rm sh}+I_{0}R_{\rm sh}-I_{\rm c}^{(i)}R_{\rm s}-I_{\rm c}^{(i)}R_{\rm sh} - \\ 
    & nV_{\rm th}\text{LambertW}\left( \frac{I_0 R_{\rm sh}}{nV_{\rm th}} \exp\left( \frac{R_{\rm sh}(-I_{\rm c}^{(i)}+I_{\rm ph}+I_{0})}{nV_{\rm th}}\right)\right)
\end{split}
\end{equation}

\begin{figure}[h] 
	\centering
	\subfloat[\label{fig:NewtonFlow}]{%
		\includegraphics[width=0.53\columnwidth]{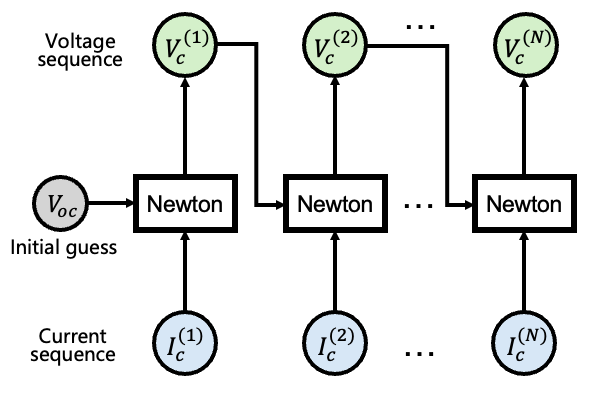}}
	\subfloat[\label{fig:LambertWFlow}]{%
		\includegraphics[width=0.47\columnwidth]{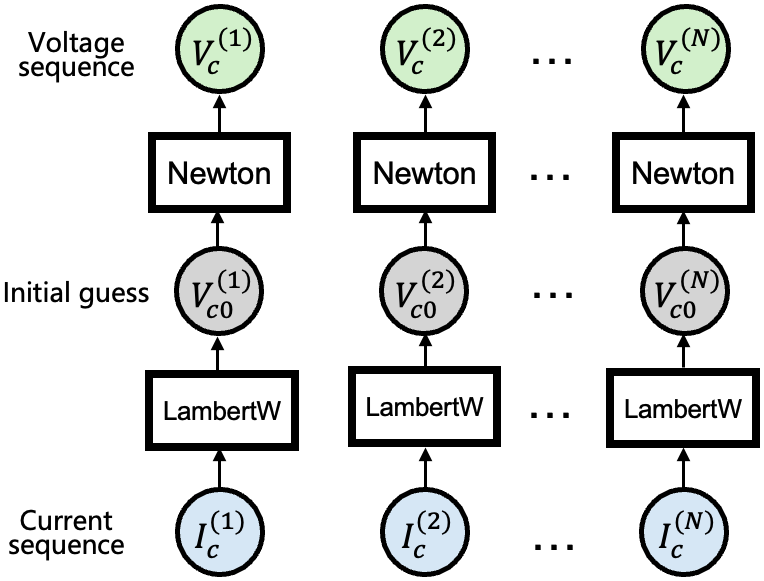}}
	\caption{The computation flow for solving RSDM using Newton method: (a) conventional recursive method, (b) our approach.}  
	\label{fig:Newton} 
\end{figure}

To demonstrate the advantages of this Lambert W function-based initial guess method, we provide Fig.~\ref{fig:Newton}, which depicts the computation flow when solving \eqref{eq-rsdm} using Newton iteration. Fig.~\ref{fig:NewtonFlow} presents a conventional way, where the initial guess for $I_{\rm c}^{(i)}$ is the voltage of the previous data point ($V_{\rm c}^{(i-1)}$) obtained from the previous Newton function. This forms a recursive way where the current calculation depends on the previous results. Such dependence reduces flexibility and increases the risk of divergence if early results incur large errors. 

In contrast, as shown in Fig.~\ref{fig:LambertWFlow}, our method independently estimates the initial guess of $I_{\rm c}^{(i)}$ using \eqref{LambertW} based on the Lambert W function, which offers the following advantages:
\begin{itemize}
    \item The initial guess for each data point is independent of previous results. This is particularly beneficial for non-ideal I-V curves, such as those that are partially missing or sparsely sampled.
    \item The initial guess is close to the true value, allowing fewer number of Newton iterations. Thus, the time complexity can be improved. 
    \item It also allows vectorized calculation, which is compatible to many tensor-based AI frameworks, e.g., PyTorch and TensorFlow. So, it can improve the implementation complexity and the computational efficiency.
\end{itemize}

Accordingly, by executing Lines 3-11 in Algorithm \ref{code:forward-calculation}, the cell voltages under different environmental conditions are obtained, denoted by the matrix $\mathbf{V_{cell}} \in \mathbb{R}^{N\times N_{\rm env}}$, corresponding to the measured current sequence $\mathbf{\tilde I} \in \mathbb{R}^N$.
\begin{equation}\label{Vcell}
\begin{split}
\mathbf{V_{cell}} = [\mathbf{v}_1, \mathbf{v}_2, \cdots, \mathbf{v}_{N_{\rm ps}}, \mathbf{v_{\rm norm}}]
\end{split}
\end{equation}
where $\mathbf{v}_i = [v_i^{(1)}, v_i^{(2)}, \dots, v_i^{(N)}]^{\top}$ represents the cell voltages under the $i$-th environmental condition. $\mathbf{v_{norm}} = [v_\text{norm}^{(1)}, v_\text{norm}^{(2)}, \dots, v_\text{norm}^{(N)}]^{\top}$ represents the cell voltages under the normal condition.

In our implementation, we fix the number of Newton iterations ($N_{\rm it}$). Consequently, Lines 3-11 in Algorithm \ref{code:forward-calculation} can be fully vectorized by eliminating the for-loops, which allows operations over arrays (or tensors) at once, rather than iterating element by element. Such vectorization yields cleaner code and improved computational efficiency.

\subsubsection{Construct $\mathbf{Q_{sub}}$ and $\mathbf{Q_{cell}}$}
Based on the designed fault vector, there are $N_{\rm env}+1$ types of substrings in the string, corresponding to $N_{\rm env}$ environmental conditions and one representing the short-circuited substring. Accordingly, the counts (number) of different substring types can be collected into a vector $\mathbf{Q_{sub}} \in \mathbb{R}^{N_{\rm env}+1}$, defined as:
\begin{equation}\label{eq:q_sub}
\mathbf{Q_{sub}} = \left[\mathbf{n_s}, N_{\rm sc}, N_{\rm sub}-N_{\rm sc}-\sum_{i=1}^{N_{\rm ps}} n_{{\rm s},i} \right]^\top
\end{equation}
where each entry represents the number of substrings of a given type; $N_{\rm sub}$ denotes the total number of substrings. 

Similarly, the number of PV cells under different environmental conditions for different substring types can be stored in a matrix $\mathbf{Q_{cell}} \in \mathbb{R}^{N_{\rm env} \times (N_{\rm env} +1)} $, defined as:
\begin{equation}\label{eq:q_cell}
\mathbf{Q_{cell}} = 
\begin{bmatrix}
n_{\rm c,1} & \cdots & 0 & 0 & 0 \\
0 & \cdots & 0 & 0 & 0 \\
\vdots & \ddots & \vdots & \vdots & \vdots\\
0 & \cdots & n_{{\rm c},N_{\rm ps}} & 0 & 0\\
N_{\rm cs}-n_{\rm c,1} & \cdots & N_{\rm cs}-n_{{\rm c},N_{\rm ps}} & 0 & N_{\rm cs}
\end{bmatrix}
\end{equation}
where each column of the matrix saves the counts of PV cells for $N_{\rm env}$ environmental conditions for a certain substring type; $N_{\rm cs}$ is the total number of PV cells in a substring.

Therefore, $\mathbf{Q_{sub}}$ and $\mathbf{Q_{cell}}$ can be considered as two structural parameters of DFFSM that encode the fault vector. 

\subsubsection{Calculate the output voltages of substrings}
The substring voltages for all substring types are represented by the matrix $\mathbf{V_{sub}} \in \mathbb{R}^{N\times (N_{\rm env}+1)}$, which is computed as:
\begin{equation}\label{Vsub}
\mathbf{V_{sub}} = \mathrm{ReLu}(\mathbf{V_{cell}} \mathbf{Q_{cell}})
\end{equation}

Here, $\mathrm{ReLu}(\cdot)$, or Rectified Linear Unit function \cite{Chen2024Domain}, is an element-wise function that outputs the entry of the matrix directly if it is positive, and outputs zero if otherwise. This is used to model the effect of bypass diodes that prevent the substring voltages from becoming negative.

\subsubsection{Calculate the output voltages of the PV string}
Finally, the voltages of the PV string are calculated as follows:
\begin{equation}\label{Vstr}
\mathbf{V} = \mathbf{V_{sub}} \mathbf{Q_{sub}} - R_{\rm c} \mathbf{\tilde I}  
\end{equation}

\begin{algorithm}[h]
\small
\caption{Forward Calculation of DFFSM}
\label{code:forward-calculation}
\begin{algorithmic}[1]  
\REQUIRE $\mathbf{\tilde I}, T, G, \mathbf{x}$ 
\ENSURE $\mathbf{V}$
\STATE $\mathbf{G} \leftarrow [(1 - r_1)G, \cdots ,(1 - r_{N_{\rm ps}})G, G]^\top$
\STATE $\mathbf{T} \leftarrow [T,\cdots ,T]^\top$
\FOR{$i = 1$ to $N_{\rm env}$} 
\STATE Calculate $I_{\rm ph}, I_{\rm 0}, n, R_{\rm s}, R_{\rm sh}$ of the PV cell
\FOR{$j = 1$ to $N$}
\STATE Calculate the initial guess of the cell voltage $V_{\rm c,0}^{(j)}$ by \eqref{LambertW}
\STATE Calculate cell voltage $v_i^{(j)}$ based on the Newton iteration method using \eqref{Vnext}
\ENDFOR
\STATE $\mathbf{v_i} \leftarrow  [v_i^{(1)}, v_i^{(2)}, \dots, v_i^{(N)}]^\top$
\ENDFOR
\STATE $\mathbf{V_{cell}} \leftarrow [\mathbf{v}_1, \mathbf{v}_2, \cdots, \mathbf{v}_{N_{\rm ps}}, \mathbf{v_{\rm norm}}]$
\STATE Construct $\mathbf{Q_{sub}}$ and $\mathbf{Q_{cell}}$ by \eqref{eq:q_sub} and \eqref{eq:q_cell}
\STATE $\mathbf{V_{sub}} \leftarrow \mathrm{ReLu}(\mathbf{V_{cell}} \mathbf{Q_{cell}})$
\STATE $\mathbf{V} \leftarrow \mathbf{V_{sub}} \mathbf{Q_{sub}}  - R_{\rm c} \mathbf{\tilde I} $
\RETURN $\mathbf{V}$
\end{algorithmic}
\end{algorithm}

\subsection{Backward Calculation of DFFSM} \label{Backward}
The backward calculation of the DFFSM computes the gradient of a designed loss function with respect to each fault parameter in the fault vector. In this paper, the loss function is defined as the mean squared error (MSE) between the measured voltage sequence ($\mathbf{\tilde V}$) from a measured I-V curve and the estimated voltage sequence ($\mathbf{V}$) from the DFFSM:
\begin{equation}\label{eq:L}
\begin{split}
    \mathcal{L}(\mathbf{x}) &= \frac{1}{N} \left\lVert \mathbf{V} - \mathbf{\tilde V} \right\rVert_2^2 \\
    &= \frac{1}{N} \sum_{i=1}^N \left( V^{(i)} - \tilde V^{(i)} \right)^2
\end{split}
\end{equation}

Denote $x$ as one fault parameter defined in the fault vector $\mathbf{x}$. The gradient of the loss function $\mathcal{L}$ with respect to $x$ is calculated by:
\begin{equation}\label{eq:dLdx}
   \frac{\rm{d} \mathcal{L}}{{\rm d} x} = \frac{2}{N} \sum_{i=1}^N \left[ \left( V^{(i)} - \tilde V^{(i)} \right) \frac{{\rm d} V^{(i)}}{{\rm d} x} \right] 
\end{equation}

Thereby, the gradient of $\mathcal{L}$ is a function of ${\rm d} V^{(i)} /{\rm d} x$, i.e., the gradient of the estimated voltage with respect to $x$, which should be derived individually for each fault parameter.
For better presentation, we remove the superscript `$(i)$' in the following derivation and assume that each voltage variable ($V$, $\tilde V$, $V_{\rm sub}$, and $V_{\rm c}$) corresponds to a common measured current value $\tilde I\in \{\tilde I^{(1)}, \tilde I^{(2)},...,\tilde I^{(N)}\}$. 

The derivation on ${\rm d} V /{\rm d} x$ for $x\in \{n_{\rm s}, n_{\rm c}, r, N_{\rm sc},  R_{\rm c}\}$ is respectively given below.

\subsubsection{${\rm d} V /{\rm d} n_{\rm s}$}
For the number of substrings affected by the $j^\text{th}$ shadow ($n_{{\rm s},j}$), where $j=1,2,\cdots,N_{\rm ps} $, based on \eqref{Vstr} the gradient is obtained by:
\begin{equation}\label{eq:dv_dns}
\frac{{\rm d} V}{{\rm d} n_{{\rm s},j}} =  V_{{\rm sub},j} - V_{\rm sub,norm}
\end{equation}
where $V_{{\rm sub},j}$ is the substring voltage under the $j^\text{th}$ shadow; $V_{\rm sub,norm}$ is the voltage of the normal substring.

\subsubsection{${\rm d} V /{\rm d} n_{\rm c}$}
For the number of shading cells per substring under the $j^{\rm th}$ shadow ($n_{{\rm c},j}$), the gradient is given by:
\begin{equation}\label{eq:dv_dnc}
\frac{{\rm d} V}{{\rm d} n_{{\rm c},j}} =  \frac{{\rm d}  V}{{\rm d} V_{{\rm sub},j}} \frac{{\rm d} V_{{\rm sub},j}}{{\rm d} n_{{\rm c},j}} = n_{{\rm s},j} \left( V_{{\rm c},j} - V_{\rm c,norm} \right)
\end{equation}
where $V_{{\rm c},j}$ is the cell voltage under the $j^{\rm th}$ shadow; $V_{\rm c,norm}$ is the voltage of the normal cell.

\subsubsection{${\rm d} V /{\rm d} r$}
For the shading ratio of the $j^{\rm th}$ shadow ($r_j$), the gradient is given by:
\begin{equation}\label{dv_dr}
    \begin{split}
        \frac{{\rm d}  V}{{\rm d} r_j} & =  \frac{{\rm d}  V}{{\rm d} V_{{\rm sub},j}} \frac{{\rm d} V_{{\rm sub},j}}{{\rm d} V_{{\rm c},j }} \frac{{\rm d} V_{{\rm c},j }}{{\rm d} r_j} 
        = n_{{\rm s},j} n_{{\rm c},j} \frac{{\rm d} V_{{\rm c},j }}{{\rm d} r_j}
    \end{split}
\end{equation}

To calculate ${\rm d} V_{{\rm c},j} /{\rm d} r_j$ in \eqref{dv_dr}, the implicit differentiation rule is applied. Because \eqref{eq-rsdm}, i.e., RSDM equation, is an implicit function of $V_{{\rm c}}$, making it challenging to get an explicit expression of $V_{\rm c}$ with respect to $r$. However, the relationship between $V_{{\rm c}}$ and $r$ is implicitly defined by $F(V_{\rm c}) = 0$ in \eqref{F}. Therefore, we have:
\begin{equation}\label{eq:dVcdr}
\begin{split}
    \frac{{\rm d} V_{{\rm c},j }}{{\rm d} r_j} & = -\frac{\partial F}{\partial r_j} \Bigl/ \frac{\partial F}{\partial V_{{\rm c},j }}
    \\
    & = -\left(\frac{\partial F}{\partial I_{{\rm ph},j}} \frac{{\rm d} I_{{\rm ph},j}}{{\rm d} r_j}
    + \frac{\partial F}{\partial R_{{\rm sh},j}} \frac{{\rm d} R_{{\rm sh},j}}{{\rm d} r_j}\right ) \Bigl/ \frac{\partial F}{\partial V_{{\rm c},j }}
\end{split}
\end{equation}
where $\partial F / \partial I_{{\rm ph},j}=1$;

\noindent ${\rm d} I_{{\rm ph},j} / {\rm d} r_j$ is:
\begin{equation}\label{eq:dIphdr}
    \frac{{\rm d} I_{{\rm ph},j}}{{\rm d} r_j}  =  -\frac{G}{G_{\rm stc}} \left[ I_{\rm ph,stc} + \alpha(T-T_{\rm stc}) \right]
\end{equation}
where $\alpha$ is the short-circuit current temperature coefficient.
$\partial F / \partial R_{{\rm sh},j}$ is:
\begin{equation}\label{dFdRsh}
    \small
    \frac{\partial F}{\partial R_{{\rm sh},j}} = 
    \frac{V_{{\rm c},j} + I_{\rm c} R_{\rm s}}{R_{{\rm sh},j}^2} + 
    \frac{a \left( I_{{\rm c}} R_{\rm s} + V_{{\rm c},j}  \right)}{R_{{\rm sh},j}^2} \left(1 -\frac{I_{{\rm c}} R_{\rm s} + V_{{\rm c},j} }{V_{\rm br}} \right)^{-m}
\end{equation}
${\rm d} R_{{\rm sh},j} / {\rm d} r_j$ is:
\begin{equation}\label{eq:dRshdr}
   \frac{{\rm d} R_{{\rm sh},j}}{{\rm d} G_j}  = \frac{G}{G_j^2}G_{\rm stc}R_{\rm sh,stc}
\end{equation}
$\partial F / \partial V_{{\rm c},j}$ is:
\begin{equation}\label{dFdVc}
\begin{split}
    \frac{\partial F}{\partial V_{{\rm c},j}} & = 
    -\frac{I_{{\rm 0}}}{nV_{\rm th}}\exp\left(\frac{V_{{\rm c},j} + I_{{\rm c}} R_{\rm s}}{nV_{\rm th}}\right) \\
    & - \frac{1}{R_{{\rm sh},j}} - \frac{a}{R_{{\rm sh},j}} \left(1 -\frac{I_{{\rm c}} R_{\rm s} + V_{{\rm c},j} }{V_{\rm br}} \right)^{-m} \\
    & - \frac{am \left( I_{{\rm c}} R_{\rm s} + V_{{\rm c},j}  \right) }{R_{{\rm sh},j} V_{\rm br}} \left(1 -\frac{I_{{\rm c}} R_{\rm s} + V_{{\rm c},j} }{V_{\rm br}} \right)^{-m-1}
\end{split}
\end{equation}

Therefore, although the forward calculation applies Newton iteration method to calculate $V_{\rm c}$, its gradient can still be obtained via the implicit differentiation rule.

\subsubsection{${\rm d} V /{\rm d} N_{\rm sc}$}
For the number of short-circuited substrings ($N_{\rm sc}$), the gradient is given by:
\begin{equation}\label{eq:dVdNsc}
  \frac{{\rm d} V}{{\rm d} N_{\rm sc}} =  -V_{\rm sub,norm}
\end{equation}

\subsubsection{${\rm d} V /{\rm d} R_{\rm c}$}
For the increased series resistance of the string ($R_{\rm c}$), the gradient is given by:
\begin{equation}\label{eq:dVdRc}
   \frac{{\rm d} V}{{\rm d} R_{\rm c}} =  -{\tilde I}
\end{equation}

To the end, the backward calculation of DFFSM provides mathematical expressions of the gradient for the loss function with respect to each fault parameter (${\rm d}\mathcal{L}/{\rm d} x$), which plays the key role in the following gradient-based fault parameters identification (GFPI) method.

\section{Gradient-based Fault Parameters Identification} \label{GradientFPI}

\subsection{Adahessian Optimizer} \label{Adahessian}
Based on the forward and backward calculations of the DFFSM introduced in Section~\ref{section-dffsm}, fault vectors can be identified using gradient-decent methods. Specifically, we adopt Adahessian \cite{Yao2021Adahessian}, a second-order stochastic optimizer that incorporates the curvature of the loss function via adaptive and fast estimates of the Hessian matrix. Compared with first-order optimizers, such as Adam or SGD, which rely solely on gradient information, Adahessian exploits second-derivative information, often achieving faster convergence and superior performance on complex optimization tasks, while maintaining computational cost comparable to first-order methods. 

Based on Adahessian, the update rule for the fault vector is formulated as follows:
\begin{equation}\label{eq:Adahessian}
    \mathbf{x}_{t} = \mathbf{x}_{t-1} - \eta_t m_t/v_t
\end{equation}
where $t=1,2,...,M$ is the iteration index, $\eta_t$ is the learning rate, and $m_t$ is the first-order moment computed as follows:
\begin{equation}\label{eq:mt}
    \begin{aligned}
        m_t = \frac{(1 - \beta_1) \sum_{i=1}^{t} \beta_1^{t-i} \mathbf{g}_i}{1 - \beta_1^t}
    \end{aligned}
\end{equation}
where $\mathbf{g}_t = {\rm{d} \mathcal{L}}/{{\rm d} {\mathbf x}_{t-1}}$ is the gradient of the loss function, and $0 < \beta_1< 1$ is the corresponding hyperparameter.

$v_t$ is the second-order moment, which is defined as:
\begin{equation}\label{eq:vt}
    \begin{aligned}
        v_t = \sqrt{\frac{(1 - \beta_2) \sum_{i=1}^{t} \beta_2^{t-i} \mathbf{D}_i^{(s)} \mathbf{D}_i^{(s)}}{1 - \beta_2^t}} 
    \end{aligned}
\end{equation}

Here, $\mathbf{D}^{(s)}$ denotes the spatially averaged Hessian diagonal (i.e.,  an averaged version of $\mathrm{diag}(\mathbf{H})$), where $\mathbf{H}$ is the Hessian matrix with elements  $h_{ij}=\partial^2\mathcal{L}/\partial x_ix_j$. $0<\beta_2<1$ is the hyperparameter for $v_t$. The detailed derivation for $\mathbf{D}^{(s)}$ from $\mathbf{H}$ can be found in \cite{Yao2021Adahessian}. 

\subsection{Projected Gradient Decent} \label{PGD}
To ensure that the fault vector is physically valid during the identification process, Adahessian update is followed by a projection step to constrain fault parameters, forming a projected gradient descent (PGD) scheme. Specifically, the unconstrained $\mathbf{x}_{t}$ produced by Adahessian update is projected onto a feasible region by:
\begin{equation}\label{eq:pgd}
    \mathbf{x}_{t}^* = \Pi_{\mathcal{C}}(\mathbf{x}_{t})
\end{equation}
where $\mathcal{C}$ denotes the feasible region for the fault vector, and $\Pi_{\mathcal{C}}(\cdot)$ is the projection operator.

The projection operator applies two types of constraints on fault parameters: individual constraints and joint constraints.

\subsubsection{Individual constraints}
First, each fault parameter will be restricted to its individual physical range via the following min-max clipping operation:
\begin{equation}\label{Constraint1}
    \mathbf{x}_{t}^* = \min(\max (\mathbf{{x}}_{t}, \bm{\ell}), \bm{u})
\end{equation}
where $\bm{\ell}$ and $\bm{u}$ denote the lower and upper bounds for the fault vector, respectively.

\subsubsection{Joint constraints}
In addition to individual bounds, some fault parameters are subject to joint constraints. For instance, the number of shaded and short-circuited substrings cannot exceed the total number of substrings. Formally, we require:
\begin{equation}\label{Constraint2}
    \sum_{i=1}^{N_{\rm ps}}n_{{\rm s},i} + N_{\rm sc} \leq N_{\rm sub} 
\end{equation}

If \eqref{Constraint2} is violated, $n_{{\rm s},i}$ will be projected to the feasible hyperplane by:
\begin{equation}\label{eq:ns_projection}
    n_{{\rm s},i}^* = n_{{\rm s},i} - \frac{\sum_{i=1}^{N_{\rm ps}}n_{{\rm s},i} + N_{\rm sc} - N_{\rm sub}}{N_{\rm env}} 
\end{equation}

Then, $N_{\rm sc}$ can be updated by:
\begin{equation}\label{eq:nsc_projection}
    N_{\rm sc}^* = N_{\rm sub} - \sum_{i=1}^{N_{\rm ps}}n_{{\rm s},i}^*
\end{equation}

Furthermore, to reduce the solution space (the number of optimal solutions) caused by different combinations of PS and the indistinguishable irradiance, a monotonic-decreasing constraint on the shading ratio $r$ across shadows is imposed, which can be formulated by:
\begin{equation}\label{Constraint4}
    r_{i+1}^* = \min(r_{i+1}, \lambda r_{i}), \,\ i = 1,\cdots,N_{\rm ps}
\end{equation}
where $0<\lambda<1$ is a predefined damping factor ensuring that the shading ratio of each subsequent shadow does not exceed a fraction $\lambda$ of that of the preceding shadow.

In summary, the pseudo-code of the GFPI is provided in Algorithm~\ref{alg:fpi}, which iteratively updates the fault vector by minimizing the discrepancy between the measured I-V curve and the DFFSM-estimated one. The algorithm initializes the fault vector and then performs $M$ optimization iterations. Each iteration consists of a forward calculation, loss evaluation, gradient computation via backward calculation, parameter updating using the Adahessian optimizer, and a projection step to enforce physical constraints. After all iterations, a final correction step $\Phi(\cdot)$ is applied to refine the results, such as rounding integer variables and discarding PS-related parameters when the shading ratio is negligible.

\begin{algorithm}[H]
\small
\caption{Gradient-based Fault Parameters Identification}
\label{code:backward-calculation}
\begin{algorithmic}[1]  
\REQUIRE $\mathbf{\tilde V},\mathbf{\tilde I},T,G$ 
\ENSURE $\mathbf{x}$
\STATE $\mathbf{x}_0 \leftarrow \mathbf{0}$  \textcolor{forestgreen}{// initial value}
\STATE $\mathbf{x}_0 \leftarrow \Pi_{\mathcal{C}}(\mathbf{x}_0)$ \textcolor{forestgreen}{// initial projection}
\FOR{$t = 1$ to $M$}
\STATE $\mathbf{ V} \leftarrow f(\mathbf{\tilde I},T,G,\mathbf{x}_{t-1})$ \textcolor{forestgreen}{// forward calculation}
\STATE $\mathcal{L}(\mathbf{x}_{t-1}) \leftarrow  {1}/{N} \left\lVert \mathbf{ V} - \mathbf{\tilde V} \right\rVert_2^2$ \textcolor{forestgreen}{// loss function}
\STATE $\mathbf{g}_t \leftarrow {\rm{d} \mathcal{L}}/{{\rm d} {\mathbf x}_{t-1}}$ by \eqref{eq:dLdx}-\eqref{eq:dVdRc} \textcolor{forestgreen}{// backward calculation}
\STATE $\mathbf{x}_{t} \leftarrow \mathbf{x}_{t-1} - \eta_t m_t/v_t$ by \eqref{eq:Adahessian}-\eqref{eq:vt} \textcolor{forestgreen}{// update by optimizer}
\STATE $\mathbf{x}_t \leftarrow \Pi_{\mathcal{C}}(\mathbf{x}_t)$ \textcolor{forestgreen}{// projection}
\ENDFOR
\STATE $\mathbf{x} \leftarrow \Phi(\mathbf{x}_{M})$ \textcolor{forestgreen}{// output correction}
\RETURN $\mathbf{x}$ 
\end{algorithmic}
\label{alg:fpi}
\end{algorithm}

\section{Experimental Validation}\label{Exp}
The proposed DFFSM and GFPI are implemented based on PyTorch framework using Python, where variables in the code are defined as differentiable tensors and most gradients defined in Section~\ref{Backward} are automatically computed via PyTorch’s autograd mechanism. We only manually specify \eqref{eq:dVcdr} (i.e., ${\rm d} V_{{\rm c}} /{\rm d} r$) as it involves implicit differentiation that cannot be recognized by Pytorch. This implementation also avoids backpropagating gradients through the Newton iteration process in the forward calculation, which prevents potential issues such as gradient explosion and vanishing. The code is open-source and available on GitHub.

The experimental validation comprises two parts. First, we use simulated I-V curves to compare the performance of different optimizers in our proposed GFPI method. Second, we assess the effectiveness of our proposed GFPI method under measured I-V curves from a real PV station.

\subsection{Comparison of Optimizers}
To demonstrate the advantages of Adahessian, this part compares its performance with five widely used gradient-based optimizers: Adam, Adamax, AdamW, AdaGrad, and RMSProp \cite{Tian2023Recent}. A total of 150 I-V curves under various fault conditions at STC are generated using a validated simulator (named CFFSM) developed in \cite{Li2019Fault}, where each curve is associated with a unique fault vector. Specifically, we applied four faults, the first PS, the second PS, SC, and SRD. Their fault parameters are listed in Table~\ref{tab:SimVector}. There are five configurations for the first PS, five for the second PS, three for SC, and two for SRD, yielding $5\times5\times3\times2=150$ combinations. 
\vspace{-5pt}
\begin{table}[h]
\centering
\caption{Fault configurations for simulated I-V curves.}
\begin{tabular}{ccc
>{\centering\arraybackslash}p{0.8cm}
>{\centering\arraybackslash}p{0.8cm}
}
\toprule
\textbf{Fault} & 1st PS & 2nd PS & SC & SRD \\
\midrule
\textbf{Parameters} &
\begin{tabular}{l}
{[0, 0, 0]} \\
{[3, 20, 0.8]} \\
{[6, 20, 0.8]} \\
{[3, 20, 0.6]} \\
{[6, 20, 0.6]}
\end{tabular}
&
\begin{tabular}{l}
{[0, 0, 0]} \\
{[3, 20, 0.4]} \\
{[6, 20, 0.4]} \\
{[3, 20, 0.2]} \\
{[6, 20, 0.2]}
\end{tabular}
&
\makecell[c]{[0]\\[6pt][3]\\[6pt][6]}
&
\makecell[c]{[0]\\[12pt][5]}
\\
\bottomrule
\end{tabular}
\label{tab:SimVector}
\end{table}

\begin{figure}[h] 
	\centering
	\subfloat[\label{fig:Loss}]{%
		\includegraphics[width=0.5\columnwidth]{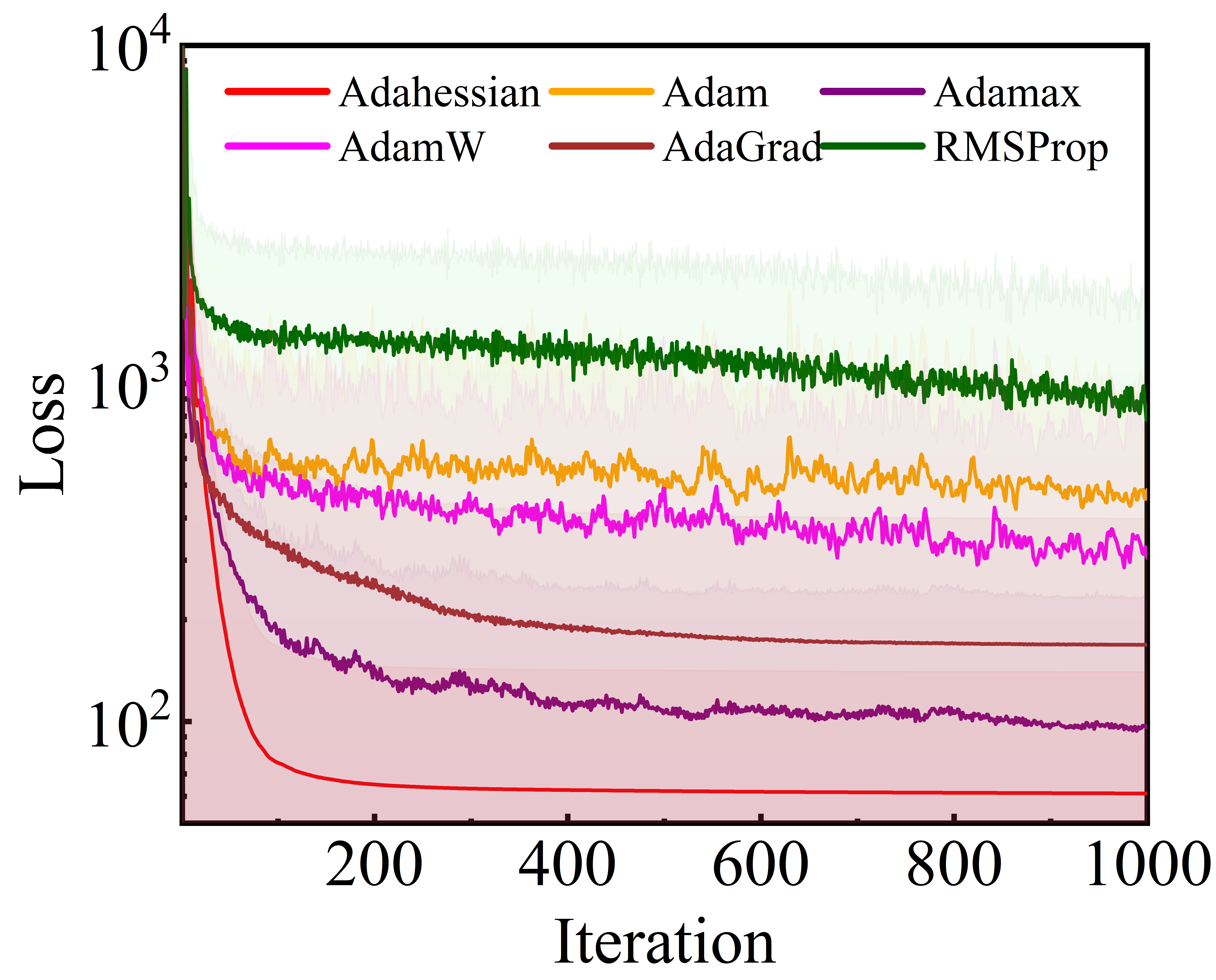}}
	\subfloat[\label{fig:Gradient}]{%
		\includegraphics[width=0.5\columnwidth]{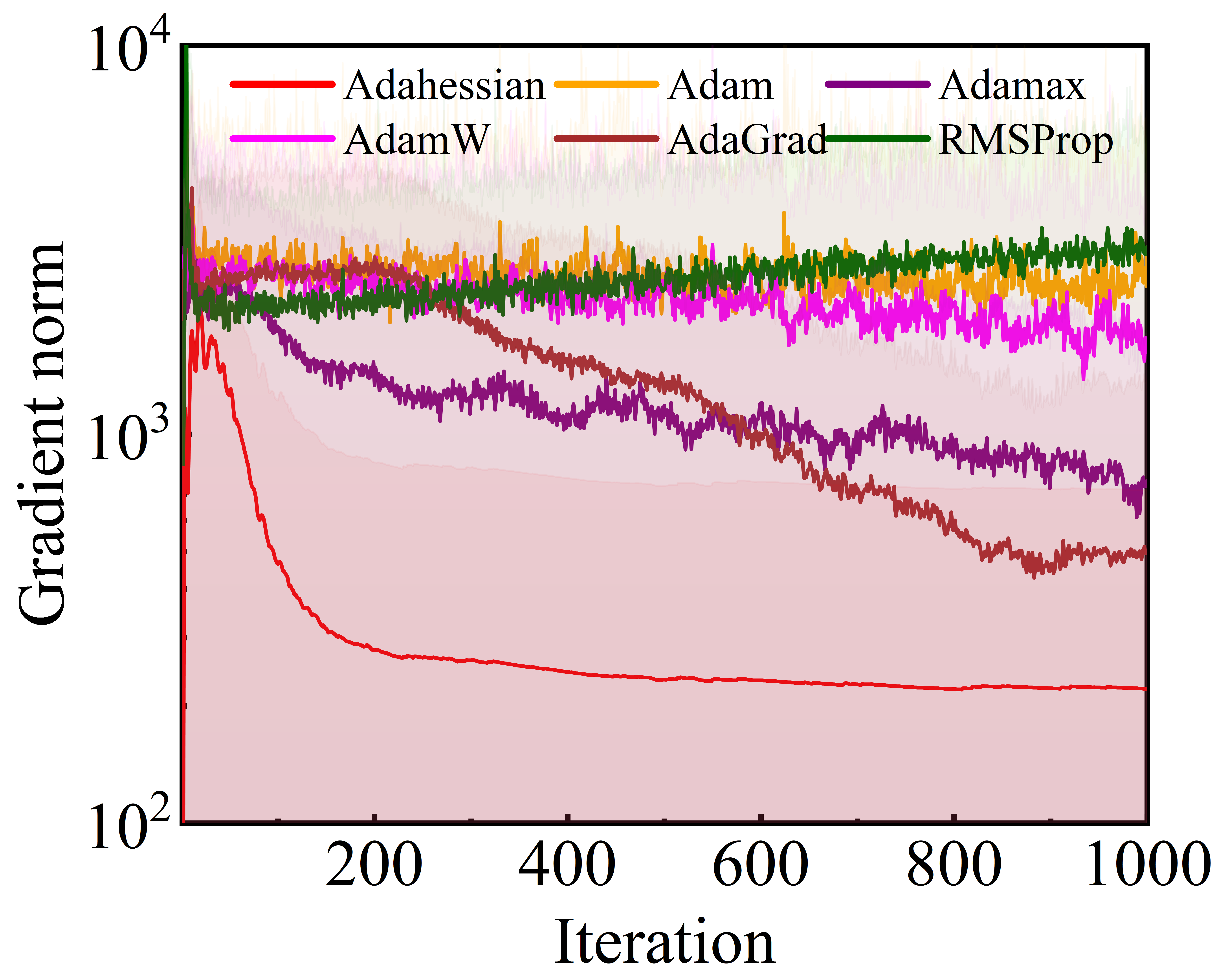}}
	\caption{Performance comparison between optimizers: (a) The average loss, (b) The average gradient norm.}
	\label{fig:LossGradient} 
\end{figure}

For every curve, we apply each optimizer independently to identify the fault parameters using 1000 iteration steps.
The performance of each optimizer is evaluated through calculating the average loss and average euclidean norm of gradient ($\|\mathbf{g}\|_2$) for each iteration over 150 I-V curves, as shown in Fig.~\ref{fig:LossGradient}. The loss, i.e., MSE, directly quantifies the identification error, while the gradient norm reflects the update magnitude of the fault vector. 

Adahessian consistently achieves the best performance across both metrics. As shown in Fig.~\ref{fig:Loss}, it rapidly enters a low-loss region within the first 200 iterations and remains stable thereafter. Ultimately, Adahessian attains the lowest average loss of 61 $\rm V^2$, whereas Adamax and AdaGrad converge to approximately 97 $\rm V^2$ and 168 $\rm V^2$, respectively. Adam, AdamW, and RMSProp exhibit slower convergence, requiring more iterations to identify fault parameters. As illustrated in Fig. \ref{fig:Gradient}, Adahessian also demonstrates faster and smoother gradient attenuation compared with other optimizers.

These advantages of Adahessian arise from its use of second-order curvature information, combined with spatial averaging and an exponential moving average to stabilize the Hessian-diagonal estimates across iterations \cite{Yao2021Adahessian}.
Therefore, Adahessian is particularly well suited for the proposed GFPI. Adamax can serve as an alternative when higher computational efficiency is required as it only requires first-order gradients.

\subsection{Experimental Validation Using Measured Data}
\begin{figure}[h]
\centering
\includegraphics[width=1\columnwidth]{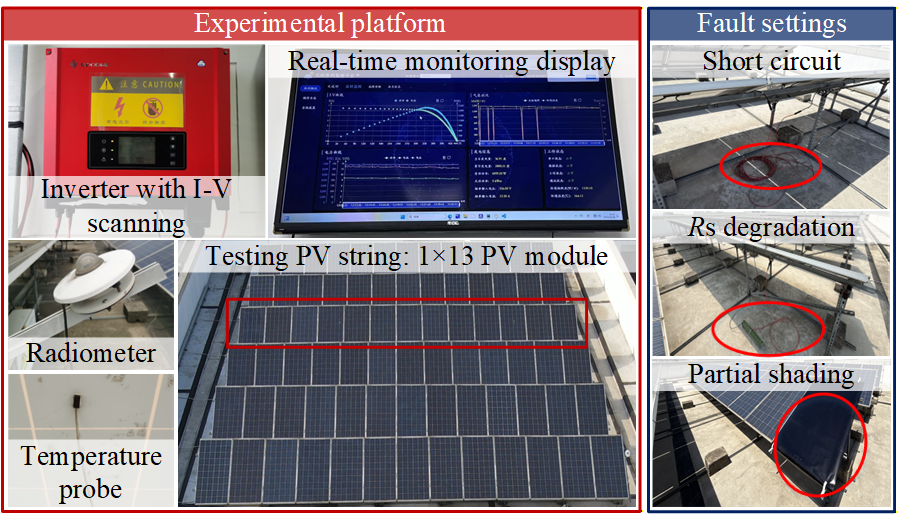}
\caption{The experimental platform and fault settings.}
\label{fig:MeaPlat}
\end{figure}

In this part, the proposed GFPI is further validated using measured data. Fig. \ref{fig:MeaPlat} shows the experimental platform and fault settings. The PV station has a 3.12kW PV string consisting of 13 240W PV modules connected in series, located in Hohai University, Changzhou Campus (31.68 $^\circ$N, 119.58 $^\circ$E). The PV modules are TSM-240 multicrystalline modules produced by Trina Solar Co., Ltd., with specifications summarized in Table~\ref{tab:tsm240}. I-V curves are directly measured by a GW20KN-DT PV inverter, a three-phase grid-connected inverter produced by GOODWE Co., Ltd. In-plane irradiance is measured by a TBQ-2 radiometer, while the module temperature is obtained by a PT100 attached to the backsheet of a PV module. SC  is implemented by short-circuiting PV modules using cables. SRD is emulated by an external resistor connected in series. PS is realized by covering PV modules with plastic films with different transmittance and areas.
\begin{table}[h]
    \centering
    \caption{Specification of TSM-240 under STC}
    \label{tab:tsm240}
    \begin{tabular}{lcc}
    \toprule
    Parameter & Symbol & Value\\
    \midrule
    Maximum power     & $P_{\rm m,stc}$ & 240 W\\
    Voltage at maximum power point  & $V_{\rm m,stc}$ & 29.7 V\\
    Current at maximum power point  & $I_{\rm m,stc}$ & 8.1 A\\
    Short-circuit current  & $I_{\rm sc}$ & 8.62 A\\
    Open-circuit voltage   & $V_{\rm oc}$ & 37.3 V\\
    Temperature coefficient of $I_{\rm sc}$ & $\alpha_{\rm stc}$ & $0.047\% / {}^{\circ}\mathrm{C}$ \\
    Temperature coefficient of $V_{\rm oc}$ & $\beta_{\rm stc}$ & $-0.32\% / {}^{\circ}\mathrm{C}$ \\ 
    Number of substrings per PV module  & $N_{\rm d}$ & 3 \\
    Number of PV cells per substring   & $N_{\rm cs}$ & 20 \\
    \bottomrule
    \end{tabular}
\end{table}

\begin{figure}[h] 
	\centering
	\subfloat[PS1\label{fig:IV_PS1}]{%
		\includegraphics[width=0.5\columnwidth]{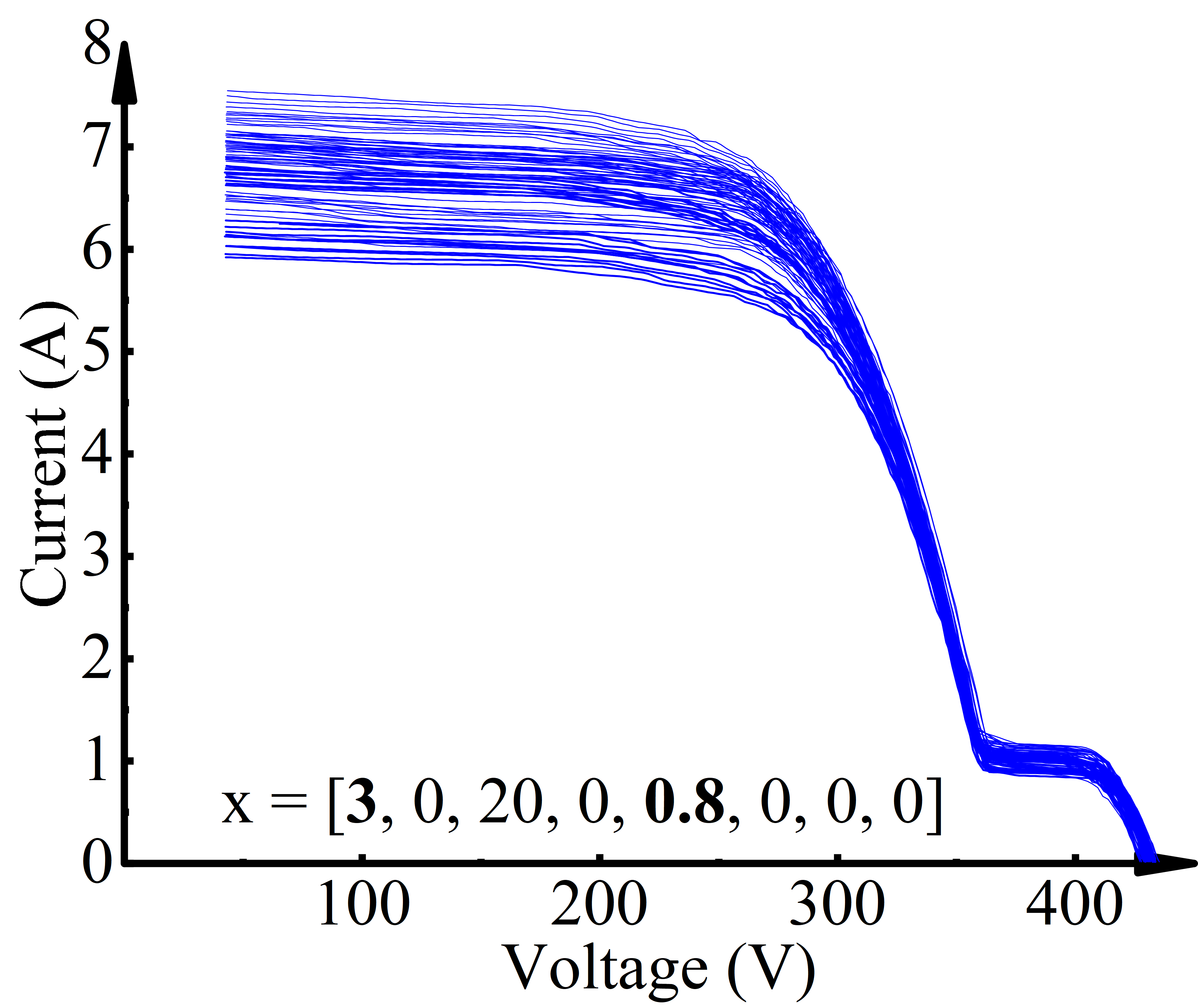}}
	\subfloat[PS2\label{fig:IV_PS2}]{%
		\includegraphics[width=0.5\columnwidth]{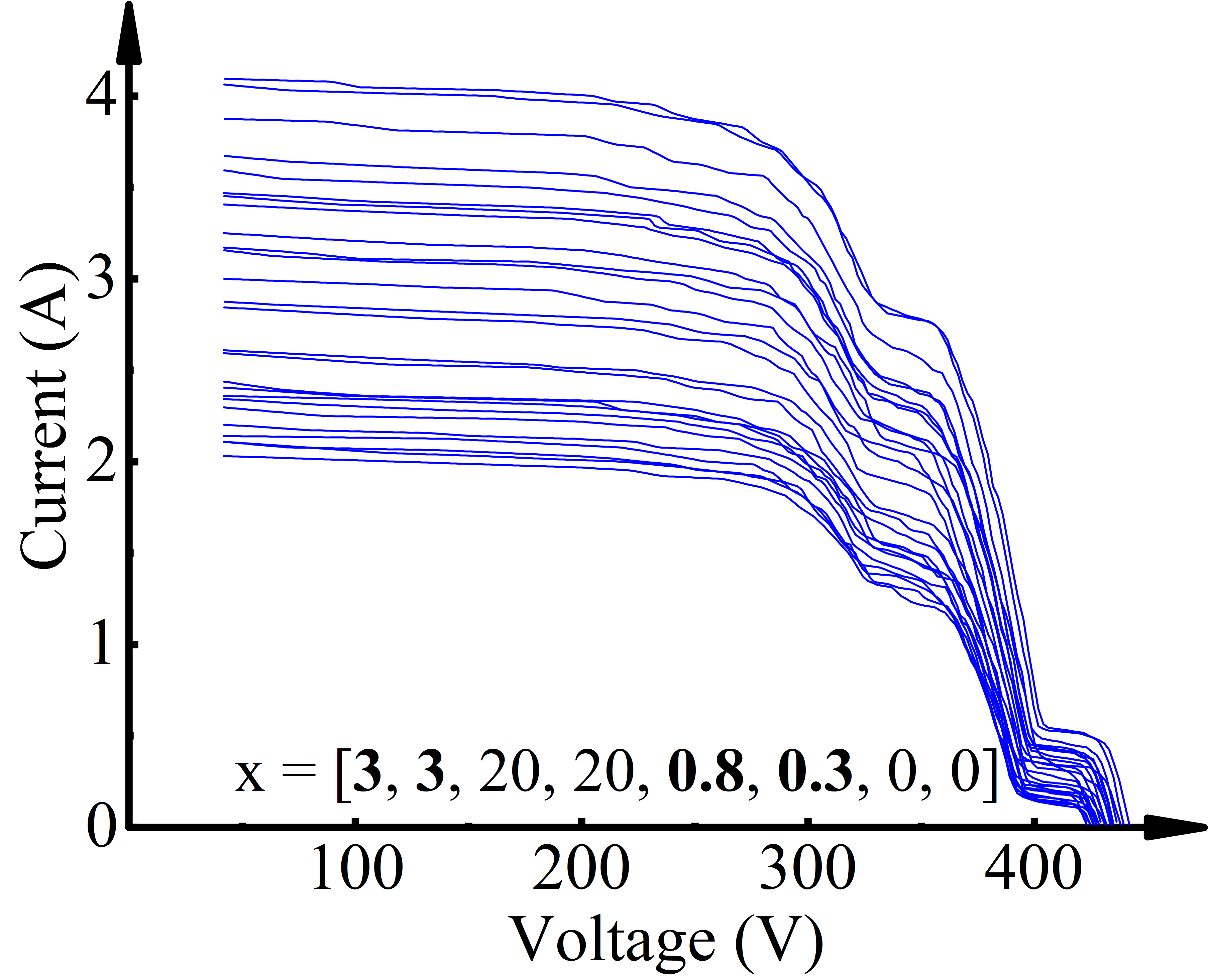}}\\
	\subfloat[SC\label{fig:IV_SC}]{%
	\includegraphics[width=0.5\columnwidth]{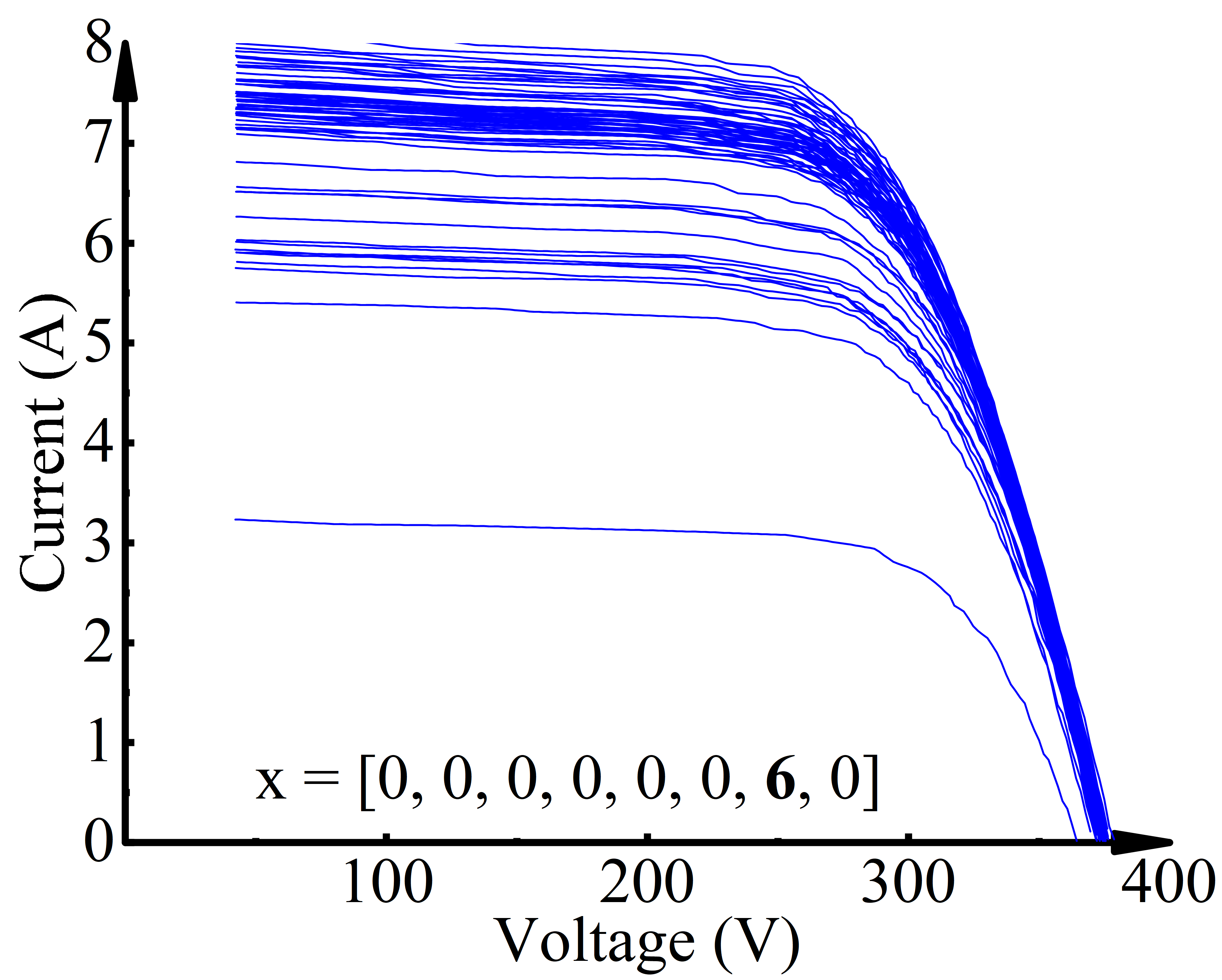}}
	\subfloat[SRD\label{fig:IV_SRD}]{%
		\includegraphics[width=0.5\columnwidth]{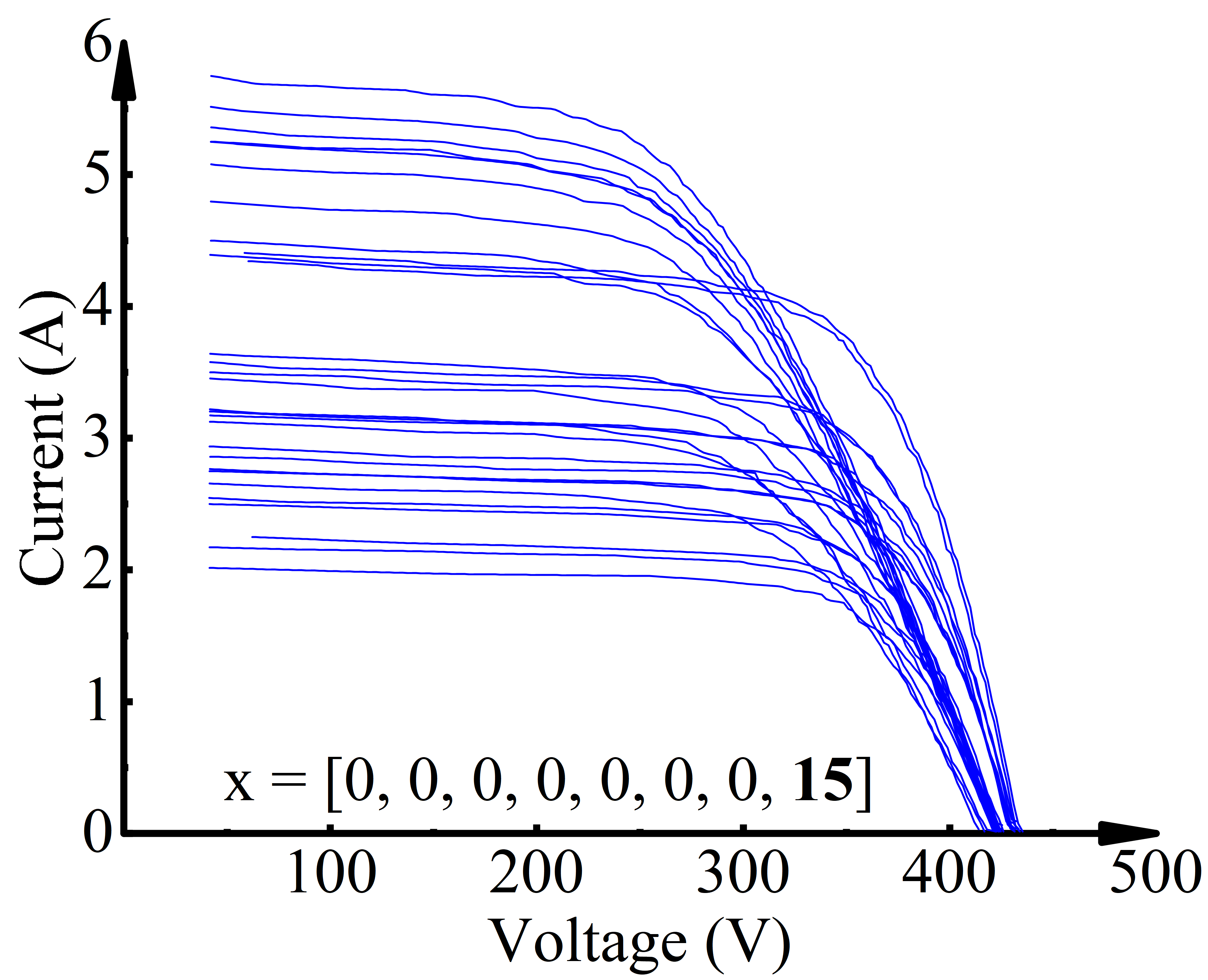}}
	\caption{Measured I-V curves with corresponding fault vectors: (a) single shading, (b) double shading, (c) SC, and (d) SRD.}  
	\label{fig:IV_Mea} 
\end{figure}

Based on this experimental platform, a total of 202 I-V curves are collected and separated into four groups, each corresponding to a specific fault vector, as shown in Fig.~\ref{fig:IV_Mea}: 83 curves under single shading, 26 under double shading, 64 under SC faults, and 29 under SRD. These groups are denoted as `PS1', `PS2', `SC', and `SRD', respectively.

For the measured I-V curves, preprocessing is performed to remove outliers and reduce noise. Moreover, since the flat-slope region between the short-circuit point and the maximum power point of an I-V curve has a disproportionately large effect on the loss value (MSE), it weakens the relative contribution of fault-induced variations in other regions of the curve. To address this issue, the data in this region are downsampled so that the optimizer focuses more on reducing errors in fault-sensitive regions.

\begin{figure}[h] 
	\centering
	\subfloat[PS1\label{fig:case_PS1}]{%
		\includegraphics[width=0.5\columnwidth]{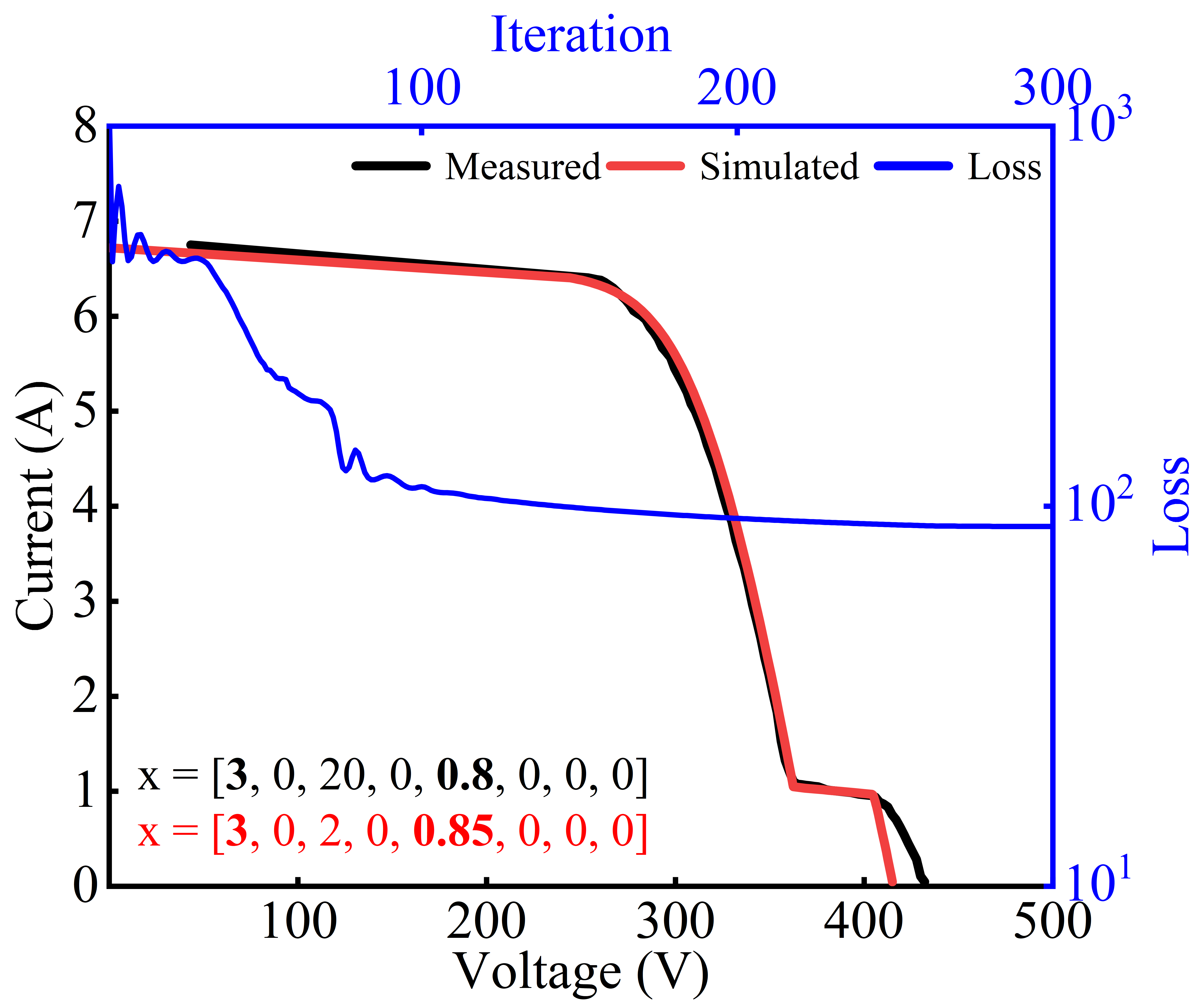}}
	\subfloat[PS2\label{fig:case_PS2}]{%
		\includegraphics[width=0.5\columnwidth]{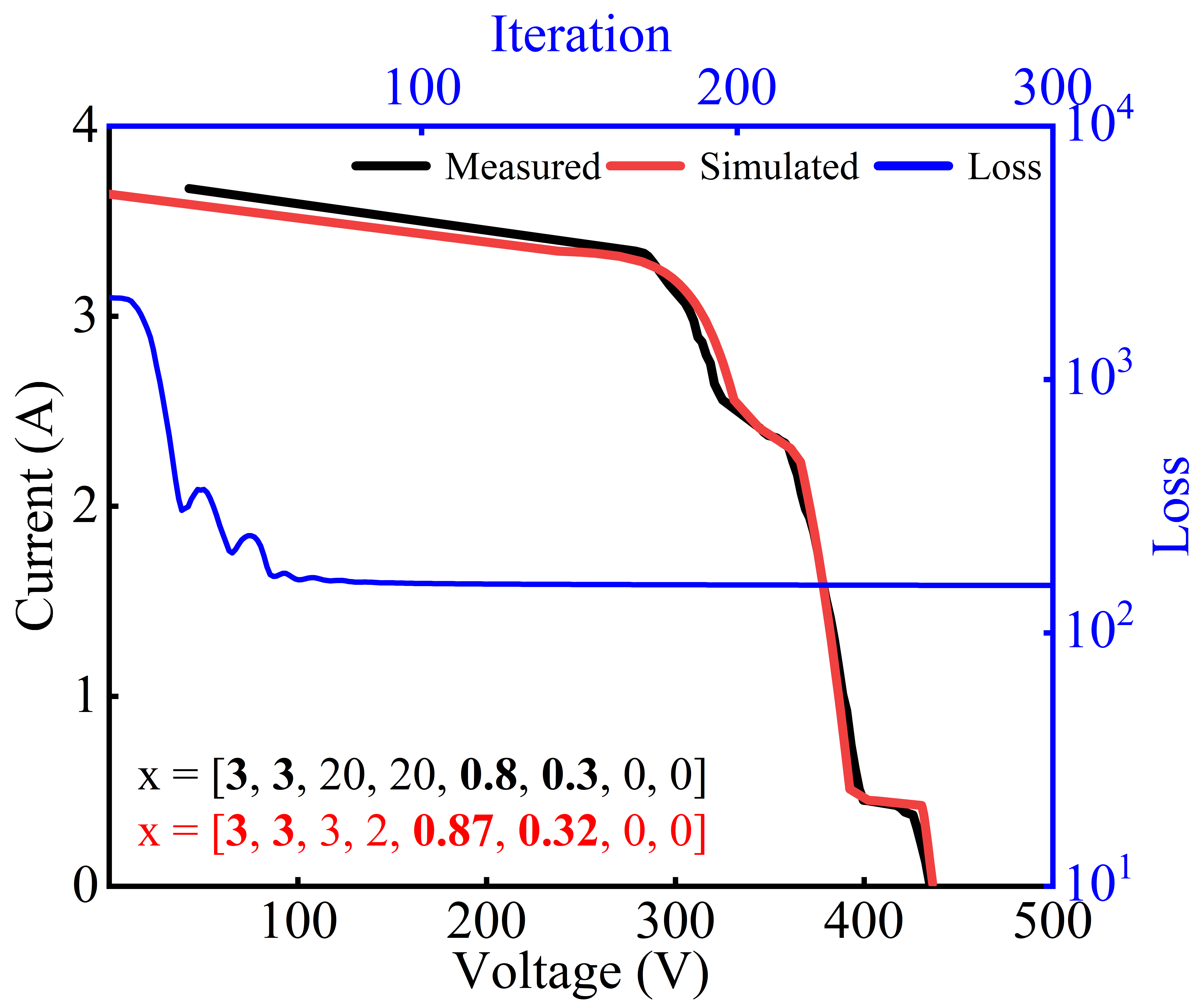}}\\
	\subfloat[SC\label{fig:case_SC}]{%
	\includegraphics[width=0.5\columnwidth]{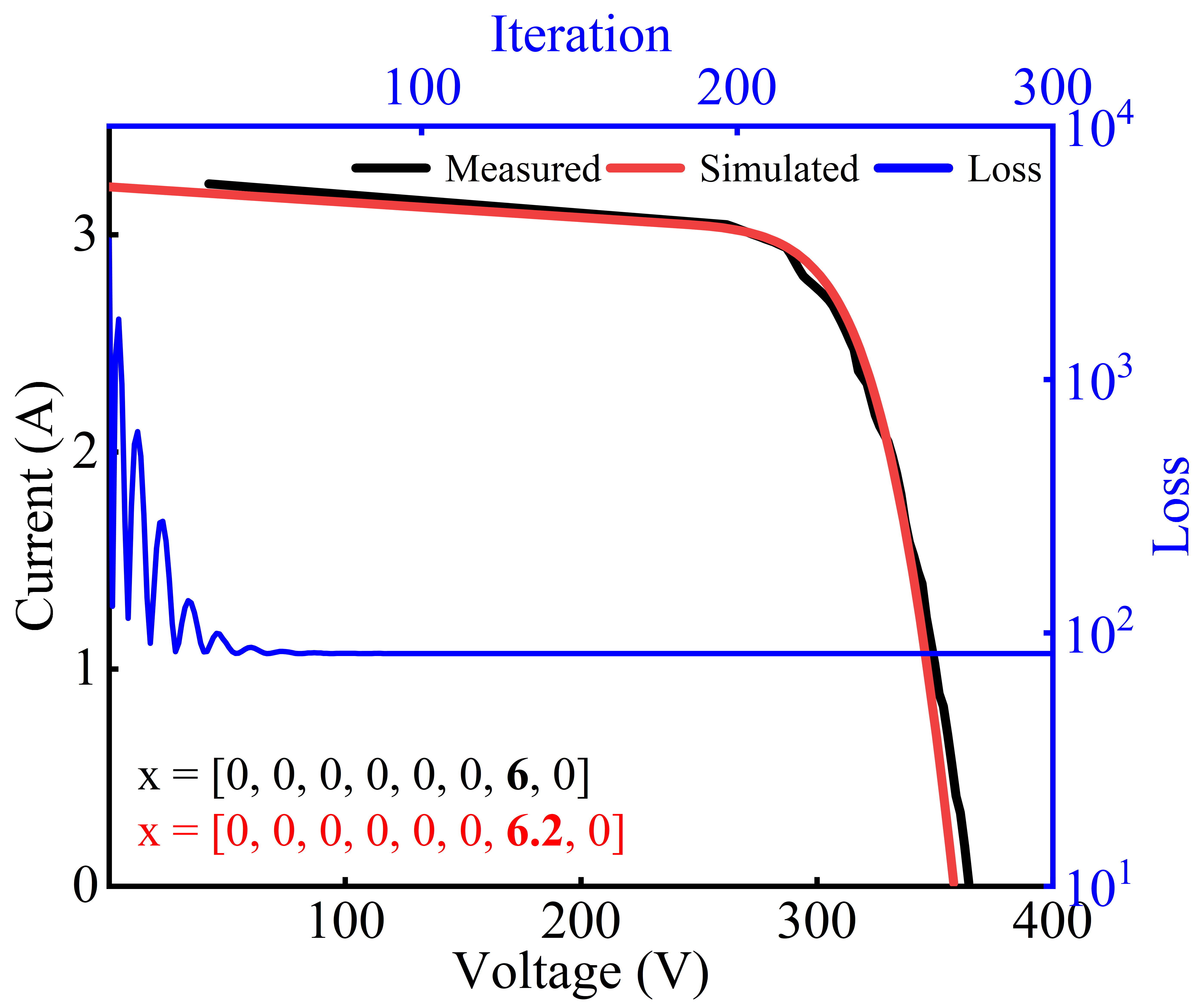}}
	\subfloat[SRD\label{fig:case_SRD}]{%
		\includegraphics[width=0.5\columnwidth]{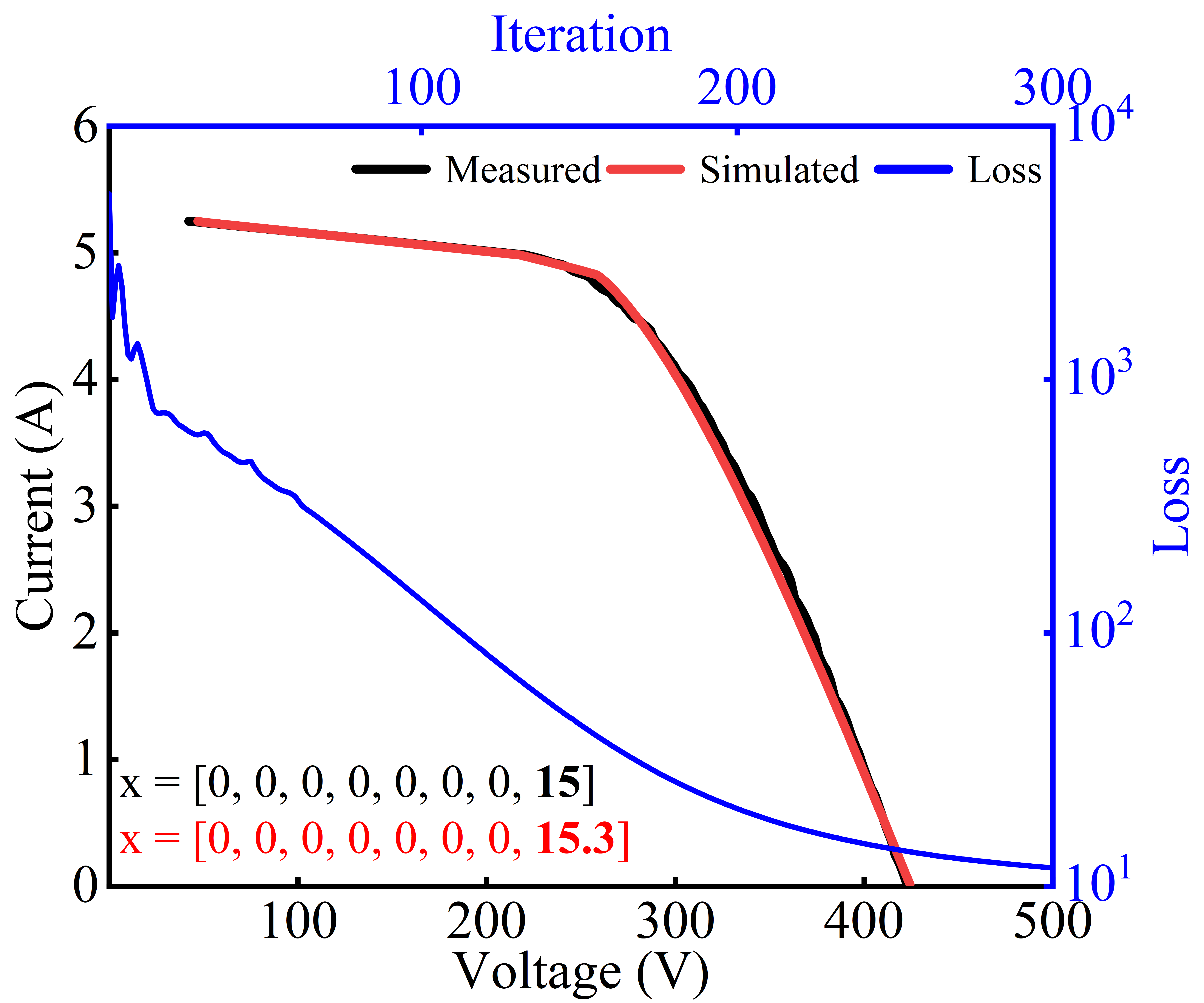}}
	\caption{Fault quantification results for cases under (a) single shading, (b) double shading, (c) SC, and (d) SRD.}  
	\label{fig:IV_Cases} 
\end{figure}

Fig. \ref{fig:IV_Cases} illustrates the quantification results for four cases, one from each group. For each case, the identified fault vector is shown together with the DFFSM-estimated I-V curve in comparison with the actual one, along with the loss curve recorded during the identification process.

As shown in Figs. \ref{fig:case_PS1} and \ref{fig:case_PS2}, the proposed GFPI method is able to identify multiple PS conditions with satisfactory accuracy. The identified number of affected substrings ($n_{\rm s}$) matches the ground truth, and the estimated shading ratio ($r$) is close to the actual value.
Nevertheless, identification errors are observed in the estimated number of shaded PV cells per substring ($n_{\rm c}$), as this parameter is less sensitive to the MSE than other parameters. This is because shading a small number of PV cells within a substring commonly creates a comparable impact to that of shading all cells.
The MSE of PS1 and PS2 cases are 88 $\rm V^2$ and 155 $\rm V^2$, respectively, corresponding to voltage errors of 2.2\% and 2.9\%.

As shown in Figs.~\ref{fig:case_SC} and \ref{fig:case_SRD}, $N_{\rm sc}$ and $R_{\rm c}$ can be precisely estimated. The loss converges rapidly for the SC case since SC have a significant impact on the voltage. The MSE of the SC case is 83 $\rm V^2$, corresponding to 2\% voltage error. The loss for SRD exhibits slower convergence but has the lowest MSE (11 $\rm V^2$), corresponding to 0.7\% voltage error.

\begin{figure}[h] 
	\centering
	\subfloat[$n_{{\rm s},1}$ \label{fig:ns1}]{%
		\includegraphics[width=0.32\columnwidth]{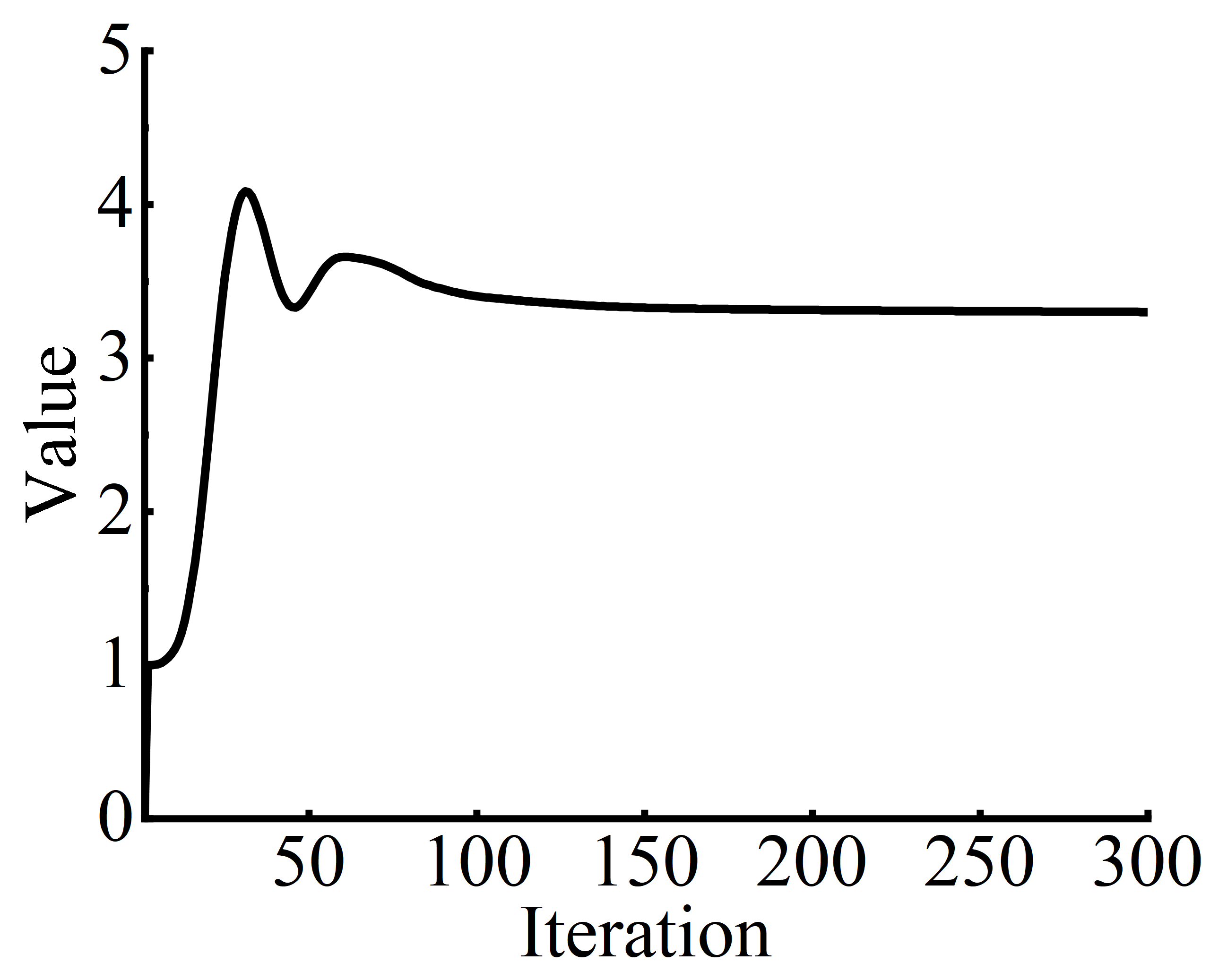}}
	\subfloat[$n_{{\rm c},1}$ \label{fig:nc1}]{%
		\includegraphics[width=0.32\columnwidth]{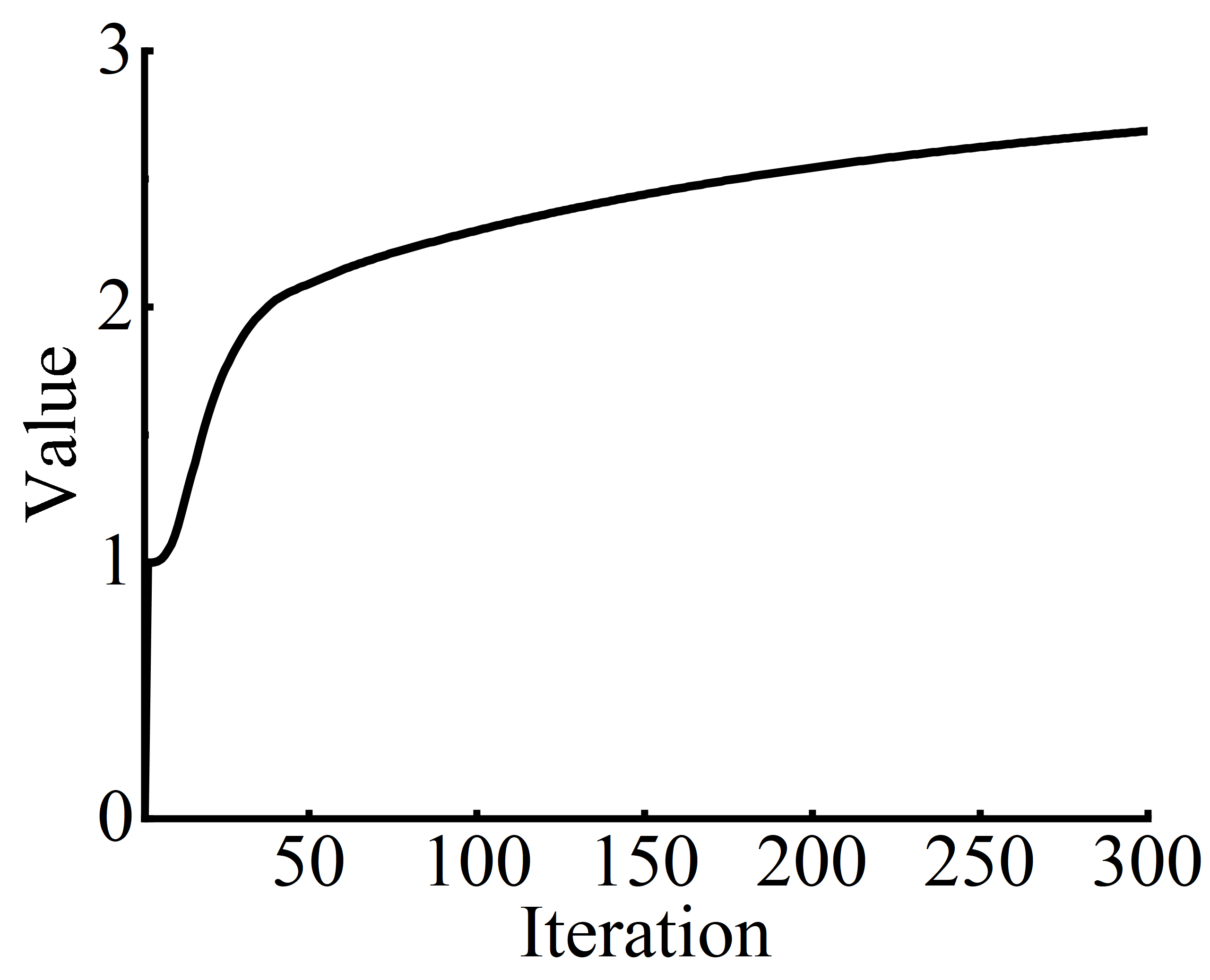}}
    \subfloat[$r_1$ \label{fig:r1}]{%
		\includegraphics[width=0.32\columnwidth]{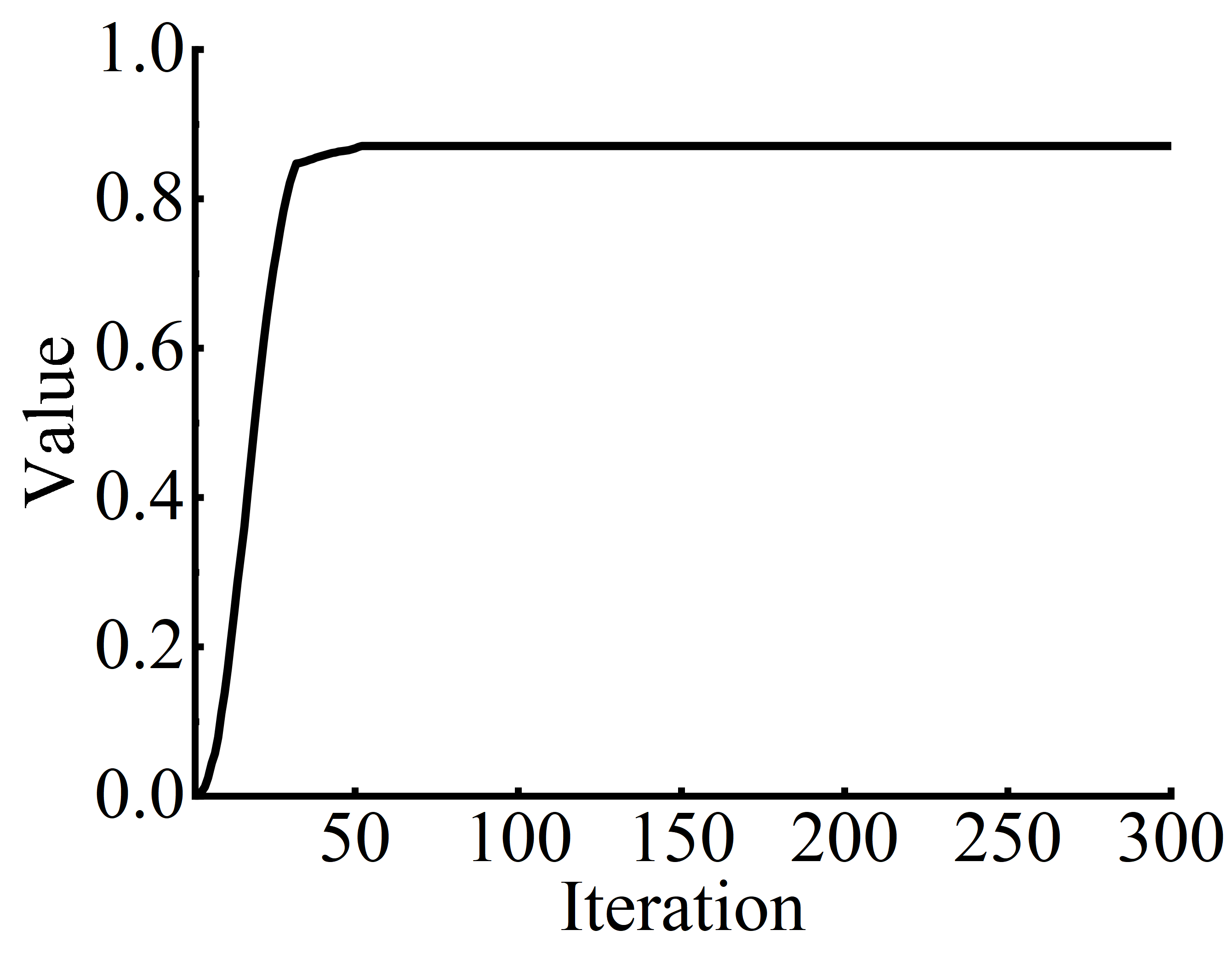}}\\
	\subfloat[$n_{{\rm s},2}$ \label{fig:ns2}]{%
	\includegraphics[width=0.32\columnwidth]{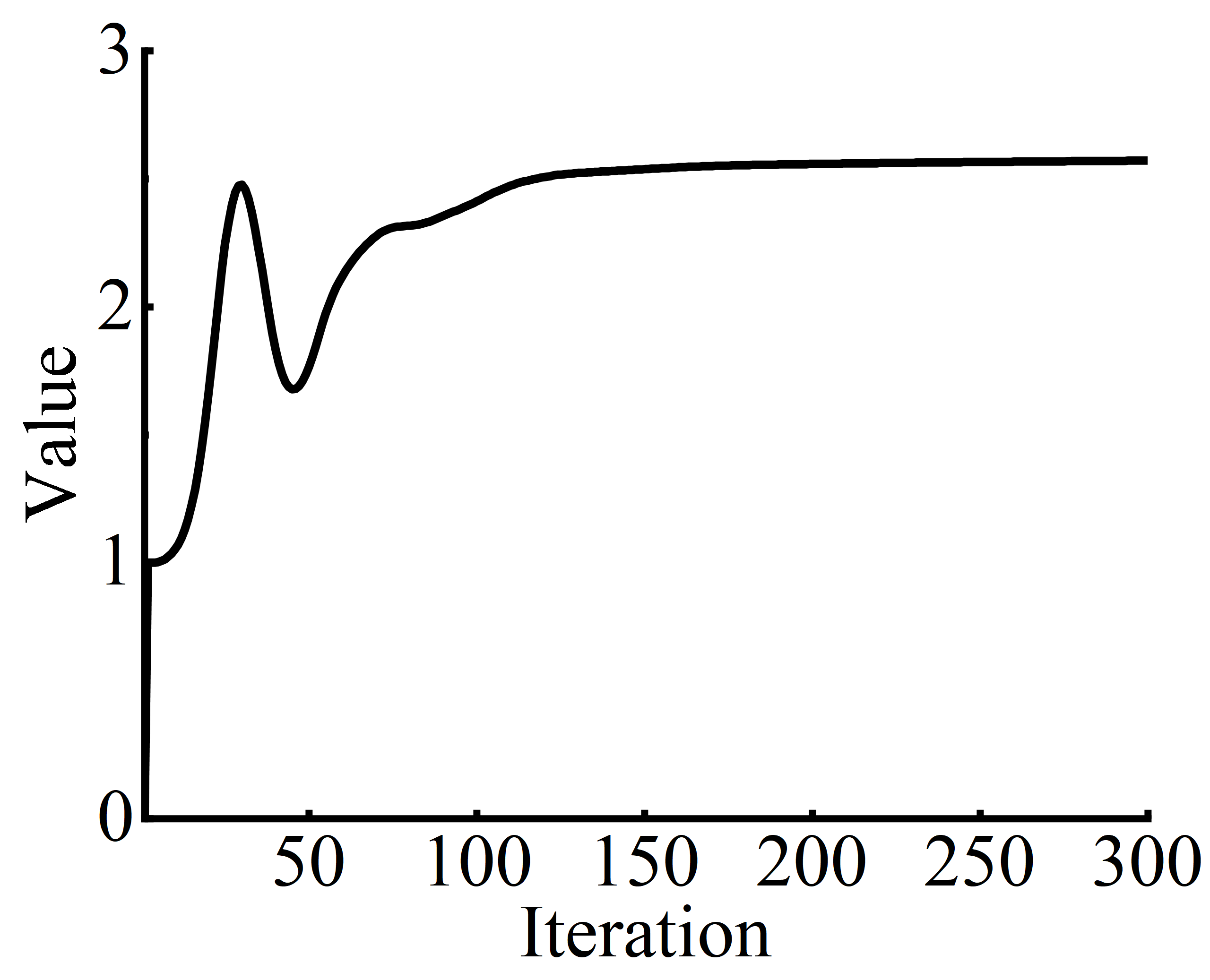}}
	\subfloat[$n_{{\rm c},2}$ \label{fig:nc2}]{%
		\includegraphics[width=0.32\columnwidth]{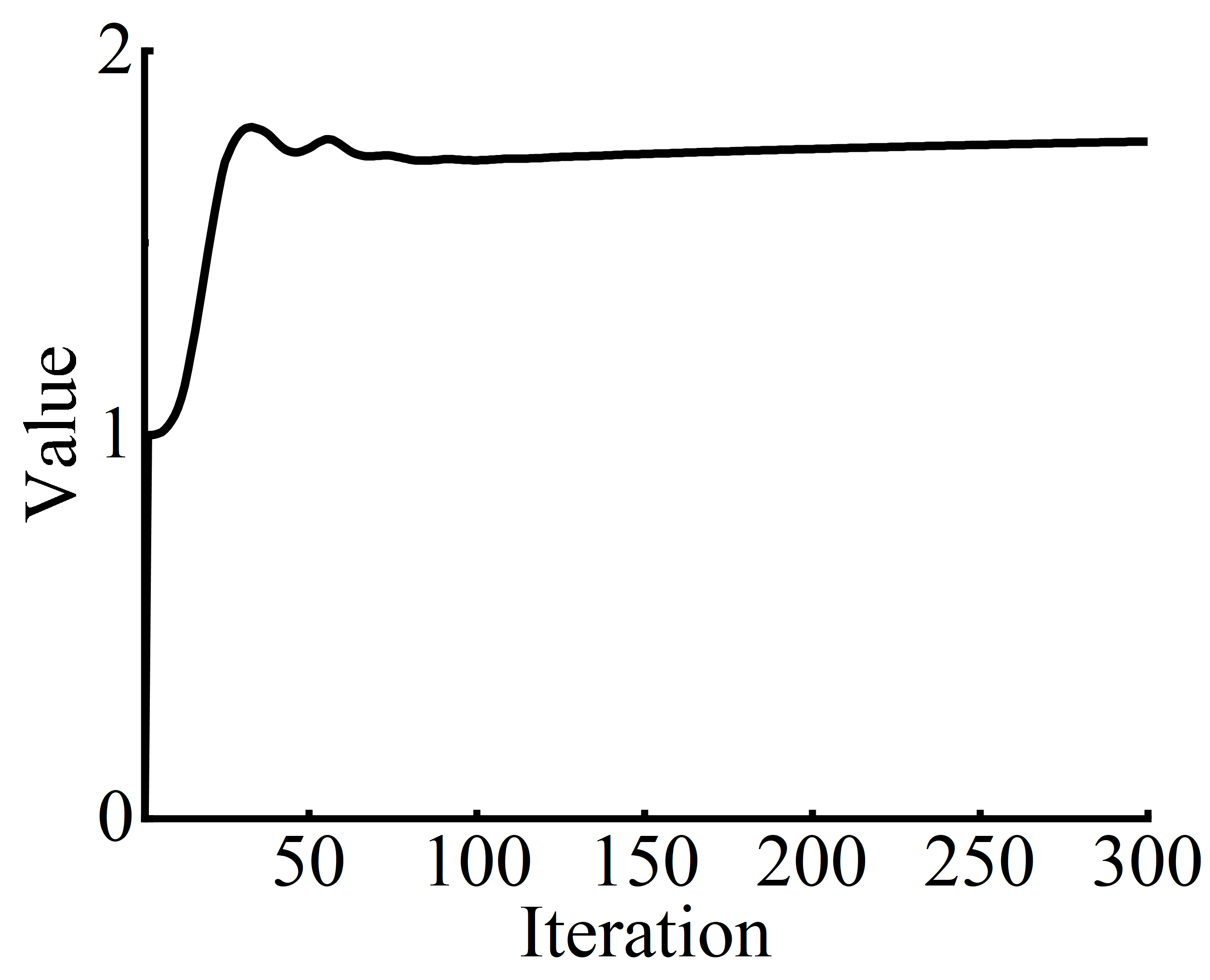}}
    \subfloat[$r_2$ \label{fig:r2}]{%
		\includegraphics[width=0.32\columnwidth]{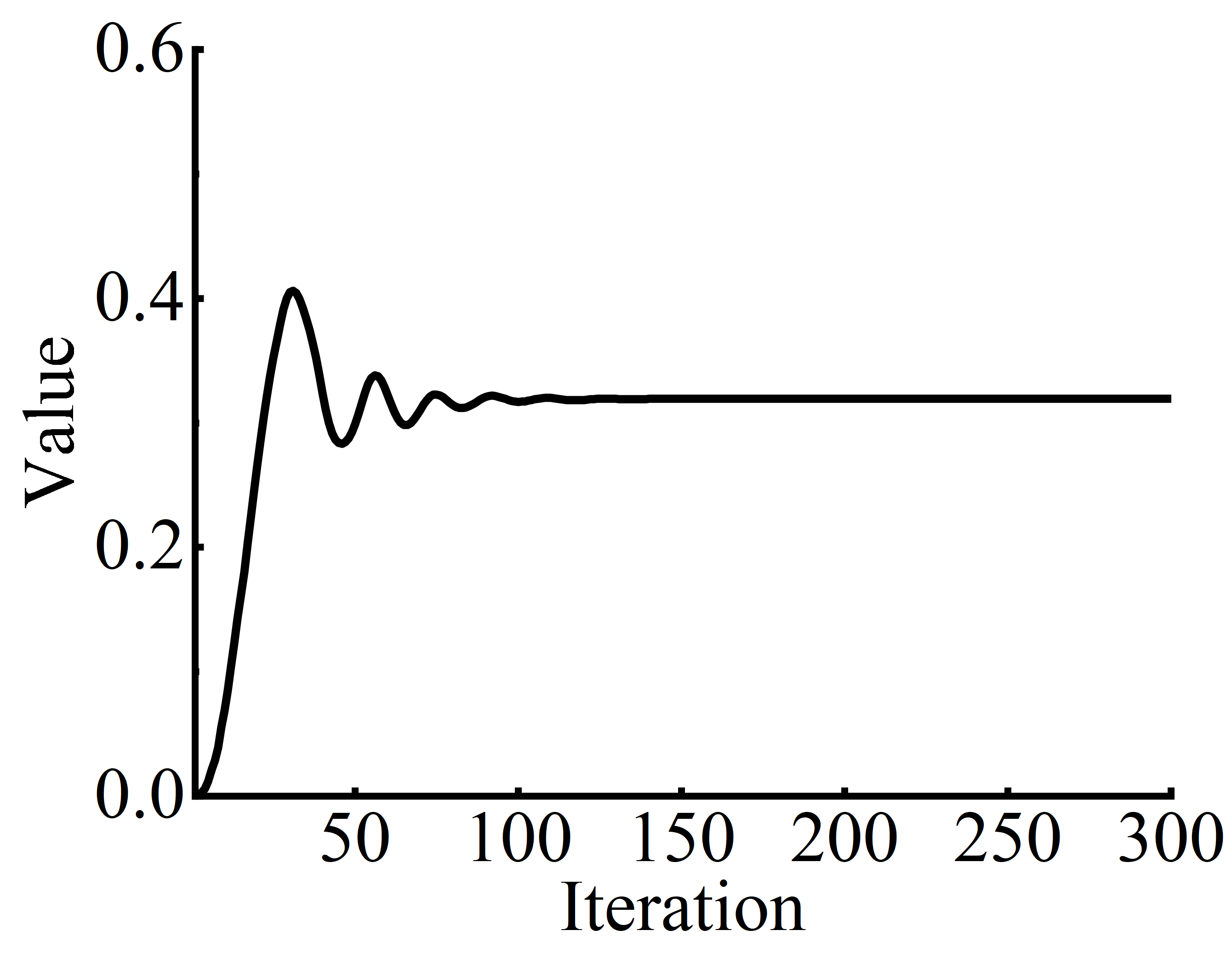}}
	\caption{Iterative evolution of the identified fault parameters for PS2 case.} 
	\label{fig:fpi} 
\end{figure}

Fig.~\ref{fig:fpi} presents the iterative evolution of the fault parameters during the identification process for the PS2 case. All fault parameters exhibit rapid adjustment in the early iterations and gradually stabilize to final values.
\vspace{-10pt}
\begin{figure}[h] 
	\centering
	\subfloat[PS1\label{fig:box_PS1}]{%
		\includegraphics[width=0.5\columnwidth]{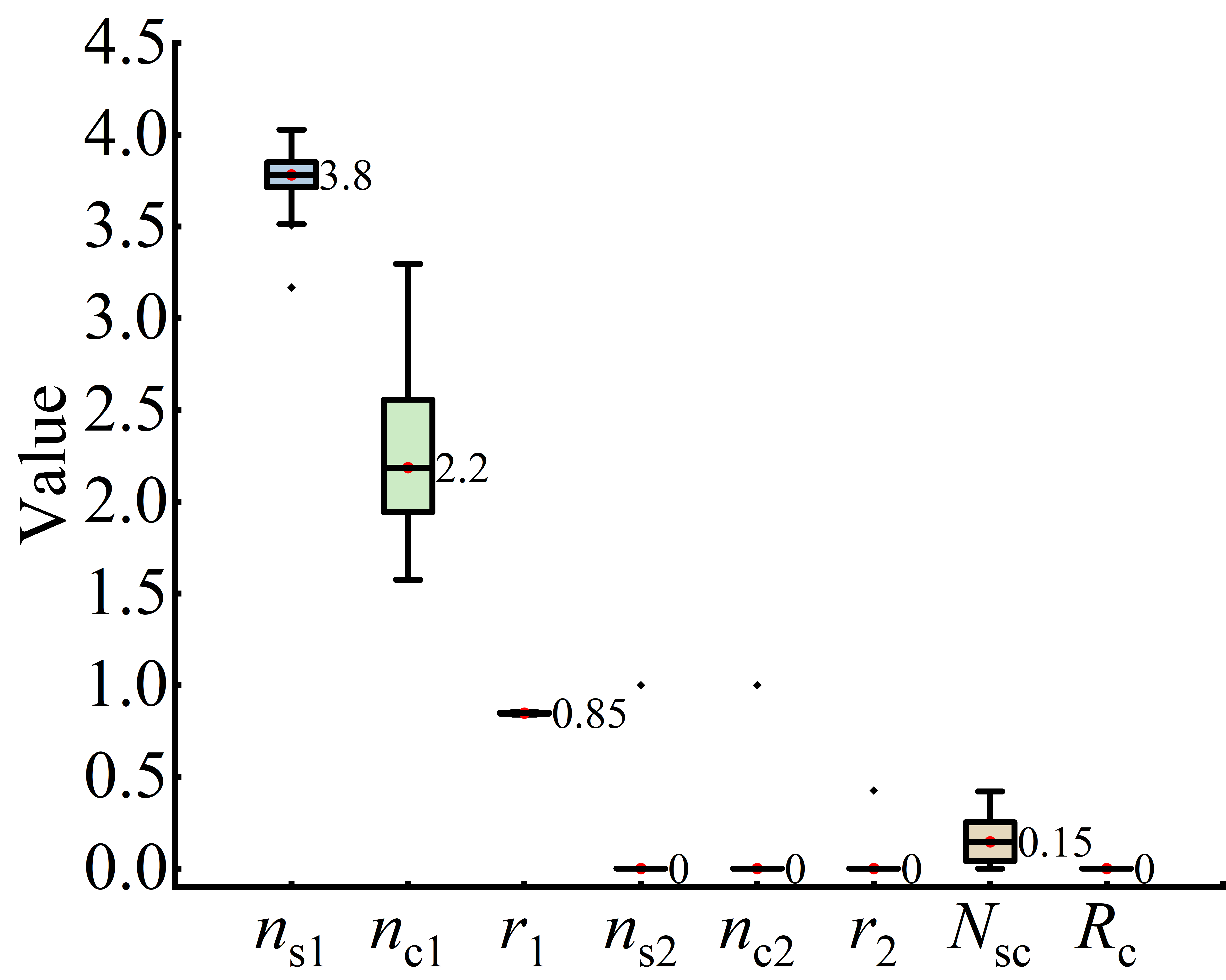}}
	\subfloat[PS2\label{fig:box_PS2}]{%
		\includegraphics[width=0.5\columnwidth]{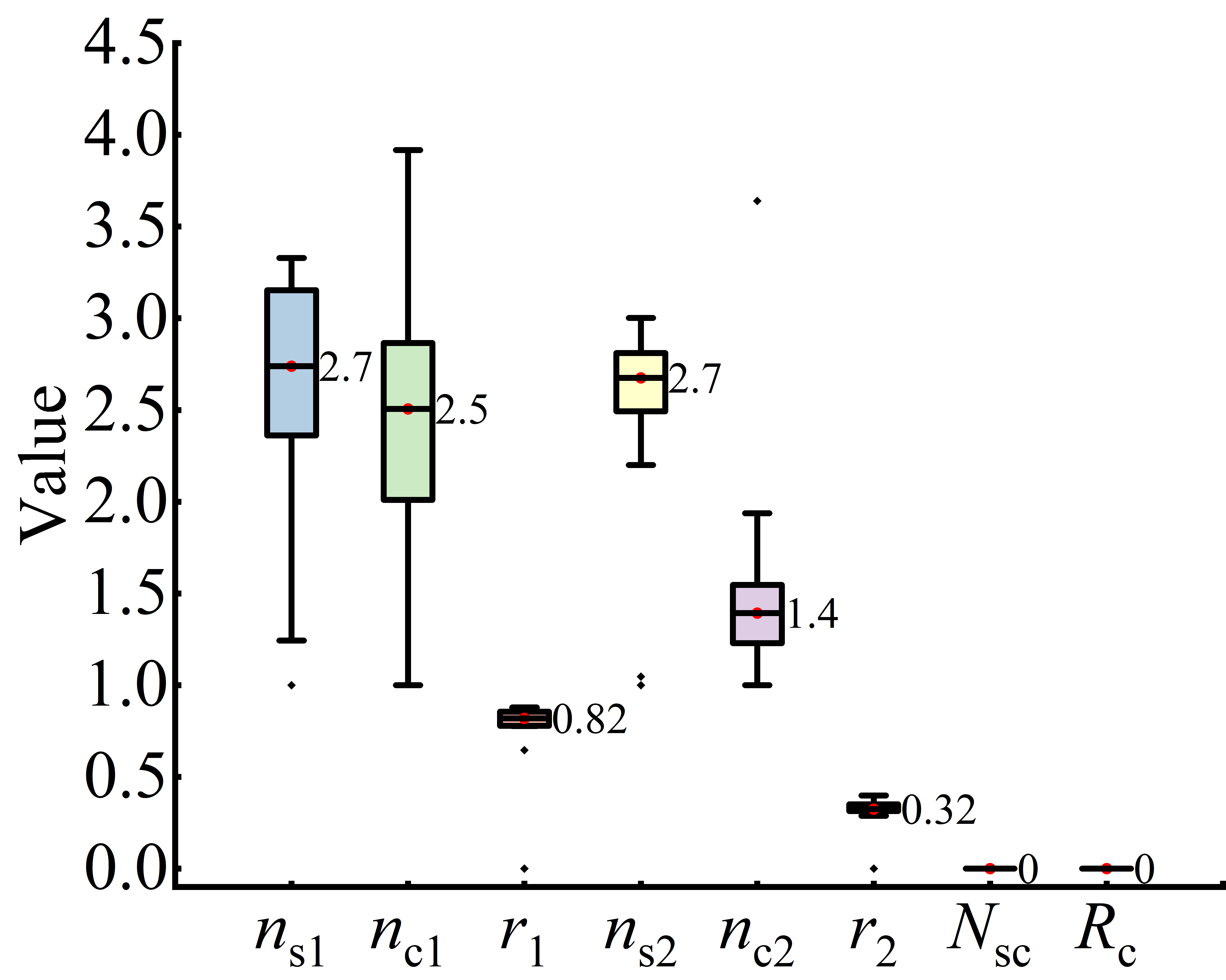}}\\
	\subfloat[SC\label{fig:box_SC}]{%
	\includegraphics[width=0.5\columnwidth]{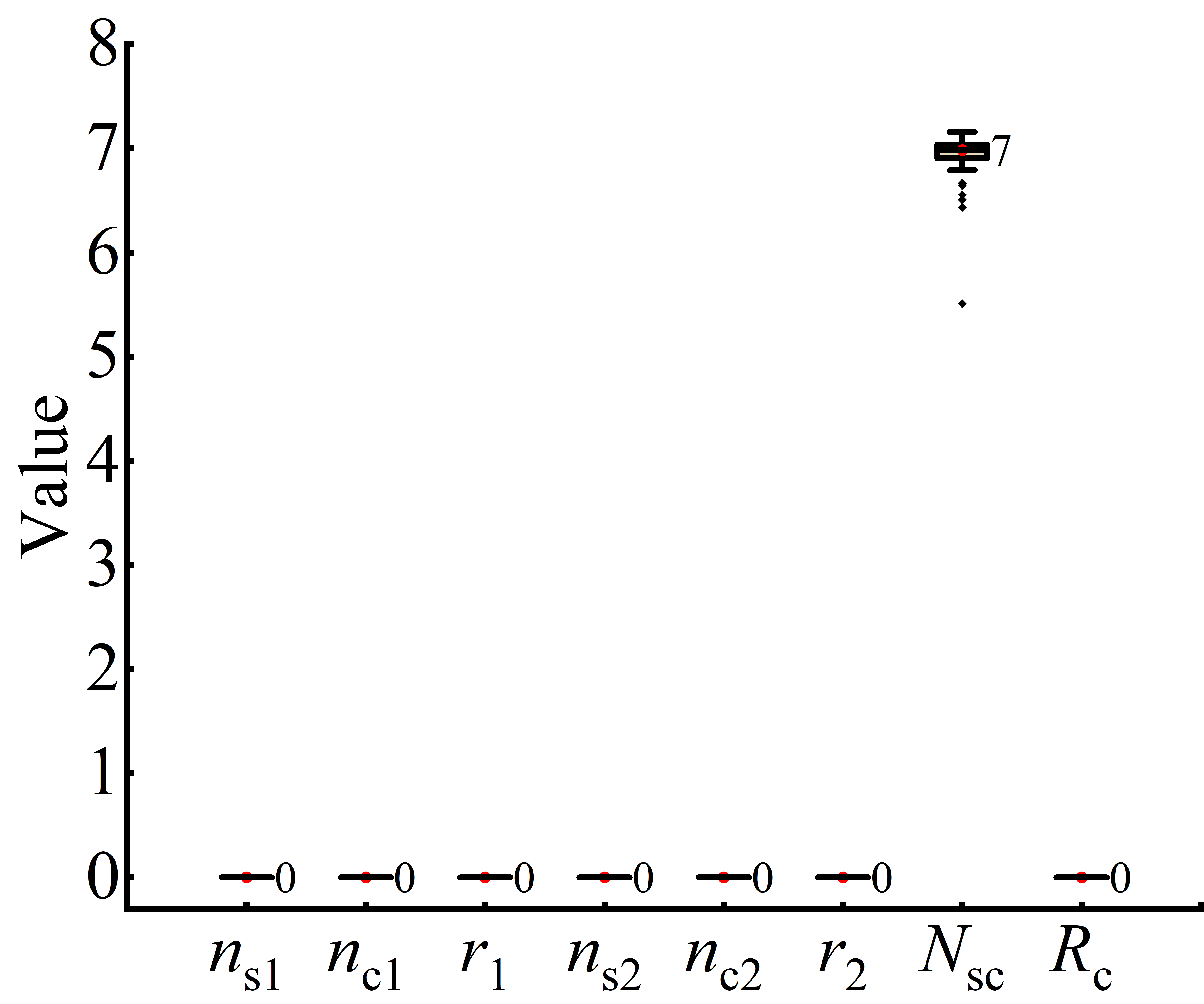}}
	\subfloat[SRD\label{fig:box_SRD}]{%
		\includegraphics[width=0.5\columnwidth]{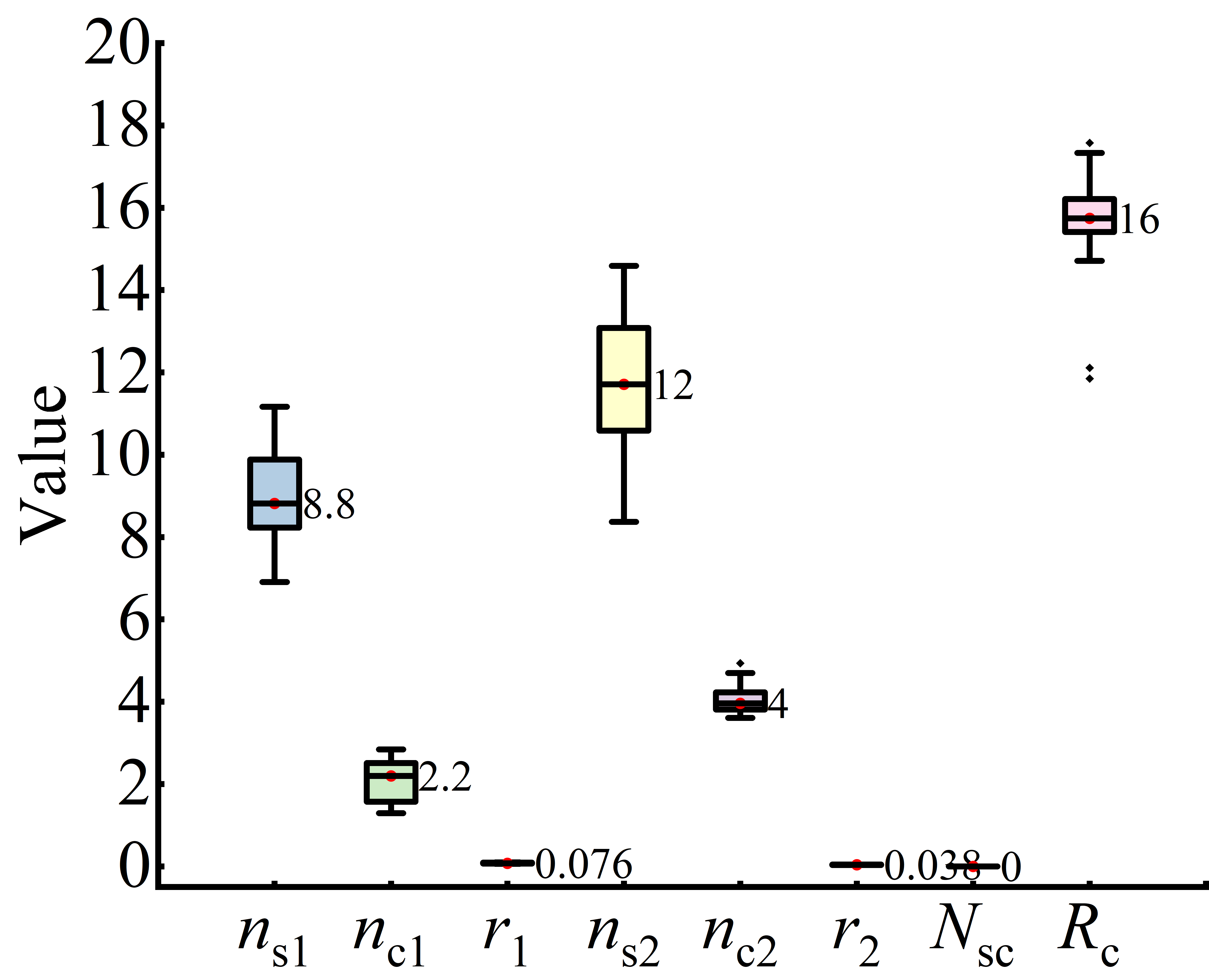}}
	\caption{Box-plots of the identified fault parameters for four groups: (a) PS1, (b) PS2, (c) SC, and (d) SRD.}  
	\label{fig:box} 
\end{figure}

Fig.~\ref{fig:box} presents box-plots of the identified fault parameters prior to the final correction step for all measured I-V curves in different groups. In the `PS1' and `PS2' groups (Figs.~\ref{fig:box_PS1} and \ref{fig:box_PS2}), shading-related parameters are accurately identified. Specifically, in the `PS1' group, the mean values of identified $n_{\rm s,1}$ and $r_1$ are 3.8 and 0.85, respectively, compared with the true values of 3 and 0.8.
Although $N_{\rm sc}$ is also identified, its value is close to zero and can be ignored after the correction step. In the `PS2' group, $n_{\rm s,1}$, $n_{\rm s,2}$, $r_1$, and $r_2$ for two shadows are all accurately identified. For the `SC' group, Fig.~\ref{fig:box_SC} shows a concentrated distribution of $N_{\rm sc}$ with a mean value of 7, which is closed to the true value of 6. For the `SRD' group, as shown in Fig.~\ref{fig:box_SRD}, the identified $R_{\rm c}$ has a mean value of 16, which is closed to the true value of 15. Although shading-related parameters are also identified, their shading ratios ($r_1$ and $r_2$) are close the zero, therefore, all shading-related parameters will be ignored after the correction step. 
\vspace{-10pt}
\begin{table}[h]
    \centering
    \caption{Overall fault quantification accuracy (RMSE)}
    \begin{tabular}{ccccc}
    \toprule
       $n_{\rm s}$ & $r$ & $N_{\rm sc}$ & $R_{\rm c}$ ($\Omega$) &
       \makecell{Reconstructed I-V\\curve (V)} \\
    \midrule
       0.75 & 0.10 & 0.69 & 1.45 & 11.07\\
    \bottomrule
    \end{tabular}
    \label{tab:Metrics}
\end{table}

Table~\ref{tab:Metrics} presents the overall accuracy of the proposed GFPI method. The root mean square error (RMSE) of the identified fault parameters is 0.75 for $n_{\rm s}$, 0.10 for $r$, 0.69 for $N_{\rm sc}$, and 1.45 for $R_{\rm c}$, respectively. The average RMSE of the DFFSM-estimated I-V curves (reconstruction error) under measured data is 11.07 V, corresponding to 2.5\% voltage error. Overall, these quantitative results further demonstrate the feasibility and effectiveness of the proposed GFPI method.

\section{Conclusions and Future Works}\label{Conclusion}
This paper develops a differentiable physical simulator, named DFFSM, for PV string fault modeling and quantification. 
By providing efficient I-V curve estimation under a given fault vector and analytical fault-related gradients, DFFSM enables a gradient-based fault parameter identification (GFPI) method proposed in this work to efficiently quantify multiple faults, including PS, SC, and SRD.
Experimental results show that the Adahessian optimizer outperforms other gradient-based optimizers under diverse simulated fault scenarios. The proposed GFPI accurately identified fault parameters from measured I-V curves for four tested fault vectors, achieving the RMSE of 0.75 for $n_{\rm s}$, 0.10 for $r$, 0.69 for $N_{\rm sc}$, and 1.45 for $R_{\rm c}$, with the I-V reconstruction errors below 3\%.

Future work will improve the DFFSM to incorporate more fault types, such as hot-spot and bypass diode open-circuit faults, and exploit its end-to-end differentiability to train AI models directly for fault quantification. Finally, we hope that the open-sourced implementation will foster more research on differentiable physical simulators and efficient fault quantification for PV systems.

\bibliographystyle{IEEEtran}
\bibliography{reference}


\end{document}